\documentclass[sigconf,anonymous=false]{acmart}
\usepackage{subfigure}
\usepackage{algorithm}
\usepackage{algpseudocode}
\usepackage{multirow}
\usepackage{threeparttable}
\usepackage{enumitem}
\usepackage{balance}
\usepackage{amsthm}

\usepackage{caption} 
\usepackage{graphicx}

\newtheorem{definition}{Definition}
\newtheorem{example}{Example}
\makeatletter
\def\old@comma{,}
\catcode`\,=13
\def,{%
  \ifmmode%
    \old@comma\discretionary{}{}{}%
  \else%
    \old@comma%
  \fi%
}
\makeatother

\makeatletter
\def\@ACM@checkaffil{
    \if@ACM@instpresent\else
    \ClassWarningNoLine{\@classname}{No institution present for an affiliation}%
    \fi
    \if@ACM@citypresent\else
    \ClassWarningNoLine{\@classname}{No city present for an affiliation}%
    \fi
    \if@ACM@countrypresent\else
        \ClassWarningNoLine{\@classname}{No country present for an affiliation}%
    \fi
}
\makeatother

\begin{document}
\title{Origin-Destination Travel Time Oracle for Map-based Services}
\author{Yan Lin$^{1,3}$, 
Huaiyu Wan$^{1,3}$, 
Jilin Hu$^2$, 
Shengnan Guo$^{1,3}$, 
Bin Yang$^2$, 
Youfang Lin$^{1,3}$,
Christian S. Jensen$^2$}
\affiliation{%
\institution{$^1$School of Computer and Information Technology, Beijing Jiaotong University, China\\
$^2$Department of Computer Science, Aalborg University, Denmark\\
$^3$Beijing Key Laboratory of Traffic Data Analysis and Mining, Beijing, China\\
}
}
\email{{ylincs, hywan, guoshn}@bjtu.edu.cn, {hujilin, byang, csj}@cs.aau.dk}

\begin{abstract}
Given an origin (O), a destination (D), and a departure time (T), an Origin-Destination (OD) travel time oracle~(ODT-Oracle) returns an estimate of the time it takes to travel from O to D when departing at T. ODT-Oracles serve important purposes in map-based services. To enable the construction of such oracles, we provide a travel-time estimation (TTE) solution that leverages historical trajectories to estimate time-varying travel times for OD pairs.

The problem is complicated by the fact that multiple historical trajectories with different travel times may connect an OD pair, while trajectories may vary from one another. To solve the problem, it is crucial to remove outlier trajectories when doing travel time estimation for future queries. 

We propose a novel, two-stage framework called Diffusion-based Origin-destination Travel Time Estimation (DOT), that solves the problem. First, DOT employs a conditioned Pixelated Trajectories (PiT) denoiser that enables building a diffusion-based PiT inference process by learning correlations between OD pairs and historical trajectories. Specifically, given an OD pair and a departure time, we aim to infer a PiT. Next, DOT encompasses a Masked Vision Transformer~(MViT) that effectively and efficiently estimates a travel time based on the inferred PiT. We report on extensive experiments on two real-world datasets that offer evidence that DOT is capable of outperforming baseline methods in terms of accuracy, scalability, and explainability.
\end{abstract}

\maketitle

\section{Introduction}
\label{sec:intro}
The diffusion of smartphones and the ongoing digitalization of societal processes combine to enable a wide range of map-based services~\cite{feng2018deepmove,guo2020attentional,lin2021pre}. Many such services rely on the availability of origin-destination~(OD) oracles that provide estimates of the travel times, distances, and paths between origin~(O) and destination~(D) locations, e.g., distance oracles~\cite{dist_oracle} and path oracles~\cite{path_oracle}. 
Examples include pricing in outsourced transportation services~\cite{cai2013fresh,wang2021passenger}, estimation of overall travel costs~\cite{litman2009transportation}, transportation scheduling~\cite{crainic1997planning,li2015towards}, delivery services~\cite{ruan2020doing,wen2021package}, and traffic flow prediction~\cite{zhang2019flow}.
For example, in flex-transport, taxi companies are paid by a public entity for making trips. The payments is based on pricing models that involve estimating the travel times of trips, but the driver is free to choose any travel path.

We study the problem of constructing an OD travel time oracle~(ODT-Oracle) that takes an OD pair and a departure time $T$ as input, and returns a travel time $\Delta t$ needed to travel from O to D when departing at time $T$. 
ODT-Oracle can provide accurate estimation of travel times that is valuable for helping public entities, taxi companies, or other transportation service providers to plan their operations more effectively, while minimizing the need for detailed travel path information.

\begin{figure}[t]
\setlength{\belowcaptionskip}{-8pt}
    \centering
    \includegraphics[width=0.9\linewidth]{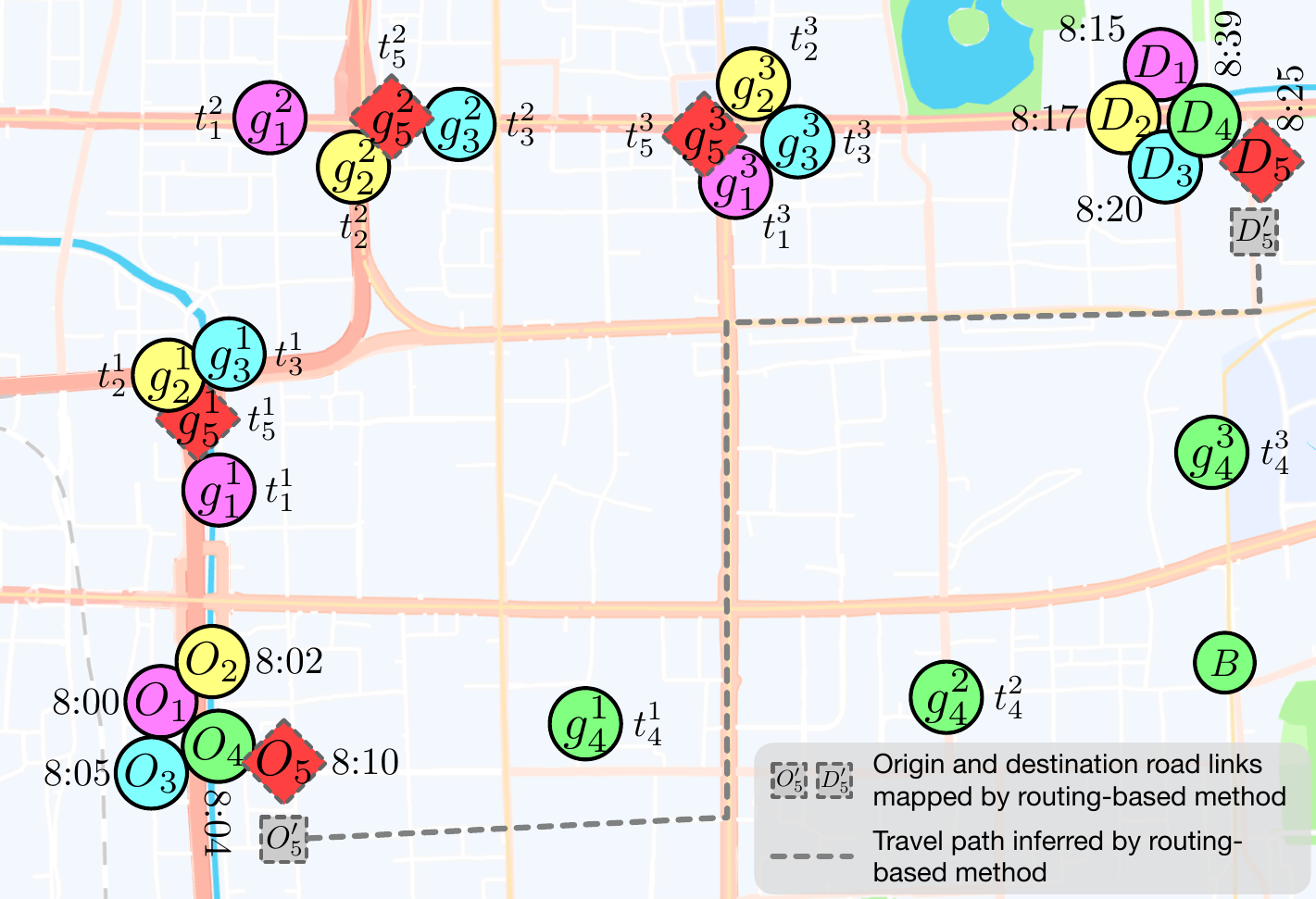}
    \caption{Motivation Example for ODT-Oracle.}
    \label{fig:odt_oracle}
\end{figure}

\begin{example}
\label{eg:outlier}
Figure~\ref{fig:odt_oracle} shows four trajectories for similar OD pair: $\mathcal T_1=\langle (O_1, \text{8:00}), (g^1_{1}, t^1_{1}), (g^2_{1}, t^2_{1}), (g^3_{1},t^3_{1}), (D_1, \text{8:15}) \rangle$, $\mathcal T_2=\langle (O_2, \text{8:02}), (g^1_{2}, t^1_{2}), (g^2_{2}, t^2_{2}), (g^3_{2},t^3_{2}), (D_2, \text{8:17}) \rangle$, $\mathcal T_3=\langle (O_3, \text{8:05}),(g^1_{3}, t^1_{3}), (g^2_{3}, t^2_{3}), (g^3_{3},t^3_{3}), (D_3, \text{8:20}) \rangle$, and $\mathcal T_4=\langle (O_4, \text{8:04}),(g^1_{4}, t^1_{4}), (g^2_{1}, t^2_{4}), (g^3_{4},t^3_{4}), (D_4, \text{8:39}) \rangle$, where each element $(g, t)$ denotes a GPS point $g$ and timestamp $t$. Thus, we have four data instances for \textbf{ODT-Oracle}: $[O_1, D_1, \text{8:00}] \rightarrow 15$, $[O_2, D_2, \text{8:02}] \rightarrow 15$, $[O_3, D_3, \text{8:05}] \rightarrow 15$, and $[O_4, D_4, \text{8:04}] \rightarrow 35$, where the travel time is the arrival time minus the departure time, e.g., $\mathcal T_1$ has travel time $\text{8:15}-\text{8:00}=15 \text{min}$. 

We observe that $\mathcal T_4$ is very different from the other three trajectories since it goes via place B, which makes it an outlier. When given a query $Q=[O_5, D_5, \text{8:10}]$, the \textbf{ODT-Oracle} is to estimate the travel time from $O_5$ to $D_5$ at 8:10. Suppose we have an unseen trajectory $\mathcal T_5=\langle (O_5, \text{8:10}), (g^1_{1}, t^1_{1}), (g^2_{1}, t^2_{1}), (g^3_{1},t^3_{1}), (D_5, \text{8:25}) \rangle$. Then, we can use the travel time of $\mathcal T_5$, $15\text{min}$, as the ground truth result for the query $Q=[O_5, D_5, \text{8:10}]$. The closer the result is to $15\text{min}$, the more accurate the \textbf{ODT-Oracle} is.
\end{example}

Studies exist that have attempted to solve this problem can be classified into two main categories~\cite{yuan2020effective}: \textbf{Non-machine learning-based method}~\cite{wang2019simple} and \textbf{machine learning-based method}~\cite{nelson1998time}. 

We can further classify \textbf{non-machine learning-based methods} into \emph{historical trajectory-based methods} and \emph{path-based methods}. TEMP~\cite{wang2019simple} is a representative \emph{historical trajectory-based method} that estimates the travel time of a given ODT-Input by averaging the travel time of historical trajectories that have a similar origin, destination and departure time. It suffers from poor accuracy when very different trajectories exist for the same O and D. For example, Figure~\ref{fig:odt_oracle} has four trajectories with similar departure times, where three similar trajectories~($\mathcal T_1$, $\mathcal T_2$, and $\mathcal T_3$) have travel time 15 minutes, while the latter ($\mathcal T_4$) goes via point B and takes 35 minutes. 
In this case, the \emph{historical trajectory-based method} returns $(15\times3+35)/4=20$ minutes, which is inaccurate due to outlier $\mathcal T_4$. 

The \emph{path-based methods} effectively solve a shortest path problem. They first map the GPS coordinates of the origin and destination onto a road network using map-matching ~\cite{Jensen2009}, i.e., $O\rightarrow O'$ and $D\rightarrow D'$. Then, they calculate the shortest path from $O'$ to $D'$ and return the travel time of this path as the estimated OD travel time. However, this type of method may also have poor accuracy due to two reasons. First, the map-matching results may be inaccurate. Second, the weights in the road network are not accurate. For example, in Figure~\ref{fig:odt_oracle}, the query origin $O_5$ and destination $D_5$, are mapped to $O_5'$ and $D_5'$, respectively. Then, a \textit{path-based method} finds the shortest path from $O_5'$ to $D_5'$, which is the dashed line in Figure~\ref{fig:odt_oracle}. However, the underlying trajectories relevant to the query $[O_5, D_5, \text{8:10}]$ do not use the shortest path, which makes the travel time estimate of \textit{path-based methods} inaccurate. 

When considering the \textbf{machine learning-based methods}, most existing studies~\cite{nelson1998time,ke2017lightgbm,jindal2017unified,li2018multi} model an \textbf{ODT-Oracle} as a regression problem without considering historical trajectories. Still, figure~\ref{fig:odt_oracle} has four historical trajectories with similar OD pairs, which are four independent training data instances for regression models.
When regression models are trained with this data by using the least squares method or mean squared error~(MSE) through backpropagation, they are likely to output 20 minutes when fed the same OD pair. Therefore, the regression-based methods also experience poor accuracy. 

The recent DeepOD~\cite{yuan2020effective} attempts to alleviate this problem by introducing an auxiliary loss. Given a historical trajectory, it first learns an OD representation with the given OD pair, departure time, and external features. It also tries to learn a representation of the given trajectory. The next step is to match the two representations, i.e., the OD and affiliated trajectory representations, which is the auxiliary loss. Then the OD representation is used to estimate the travel time that is compared with the ground truth, which is the main loss. Finally, DeepOD is trained with the combination of main loss and the auxiliary loss, which enables performance improvements. However, in the example in Figure~\ref{fig:odt_oracle}, outlier $\mathcal T_4$ is still used for training in DeepOD, which is the key reason for the reduced accuracy of existing solutions for \textbf{ODT-Oracles}.

\begin{figure}[t]
\setlength{\belowcaptionskip}{-12pt}
    \centering
    \includegraphics[width=1.0\linewidth]{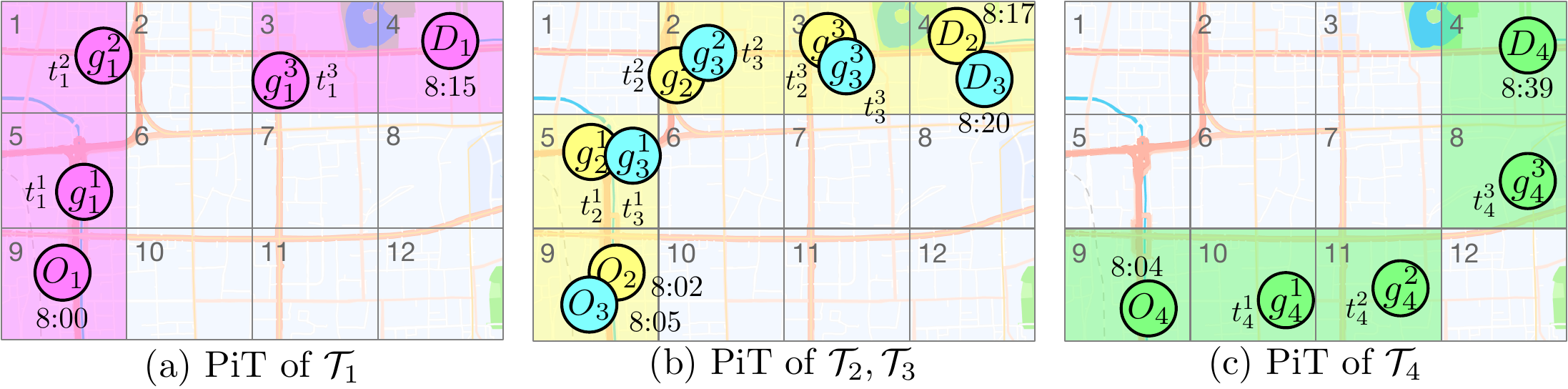}
    \caption{Examples of Pixelated Trajectories (PiTs).}
    \label{fig:pit}
\end{figure}

In this paper, we propose a new trajectory format, namely that of a Pixelated Trajectory~(PiT). We represent a PiT using an image format, which is denoted as $\mathbb R^{N\times M \times C}$, where $N$ and $M$ are the height and width of the PiT, respectively, and $C$ denotes the number of features in the PiT, which consists of a Mask, a Time of the day~(ToD), and a Time offset. 
Figure~\ref{fig:pit} shows examples of PiTs for $\mathcal T_{1}$, $\mathcal T_{2}$, $\mathcal T_{3}$, and $\mathcal T_{4}$. In the example, we use $N=3$, $M=4$, and $C=1$, so the PiT has the format $X\in \mathbb R^{3\times 4 \times 1}$, where $X[\cdot, \cdot, 1]$ is the mask that captures whether the cells are traversed by the trajectory or not. 
Although $\mathcal T_{1}$, $\mathcal T_{2}$, and $\mathcal T_{3}$ are different, their corresponding PiTs are very similar. This similarity helps the model learn common patterns and characteristics from them. On the other hand, $\mathrm{PiT}_4$ is quite different from the other PiTs, which allows for easier identification and removal of this PiT as an outlier. This demonstrates the effectiveness of the proposed PiT representation in handling different trajectories and identifying outliers.

To remove the impact of outliers, we propose a novel, two-stage ODT-Oracle framework, called \textit{\underline{D}iffusion-based \underline{O}rigin-destination \underline{T}ravel Time Estimation}~(DOT).

In the first stage, we propose a \textit{PiT inference model} to infer the PiT for a query $(O, D, T)$. Given the example in Figure~\ref{fig:odt_oracle} and the query $Q=[O_5, D_5, \text{8:10}]$, we aim to generate a PiT that is similar to $PiT_{1}$, $PiT_{2}$, and $PiT_{3}$, but is different from $PiT_{4}$. 
To do this, we first propose a conditional diffusion model to generate a PiT based on the origin, destination, and departure time. Then, we leverage historical trajectory data to train this model such that it can generate a PiT when $(O, D, T)$ is given.

In the second stage, called \textit{PiT travel time estimation}, we aim to estimate travel times based on the generated PiTs. For example, given the PiTs in Figure~\ref{fig:pit}, we generate a   PiT that is similar to $PiT_1$, $PiT_2$, and $PiT_3$ but is different from $PiT_4$. So, given the query $Q=[O_5, D_5, \text{8:10}]$, the outlier $\mathcal T_4$ is disregarded during estimation. 

Since the inferred PiT is formed by cells with spatial-temporal features, we attempt to model the global correlations in PiTs based on self-attention and the vision Transformer~(ViT)~\cite{DBLP:conf/iclr/DosovitskiyB0WZ21} to improve the estimation accuracy. 
Considering that a PiT only occupies very few cells in the whole image, we further propose a Masked Vision Transformer~(MViT) equipped with an efficient self-attention masking scheme to speed up the estimation process. The efficiency of training and estimation are improved substantially compared to the original ViT. 

In summary, we make the following contributions:
\begin{enumerate}[leftmargin=*]
    \item We propose a novel two-stage framework for enabling accurate ODT-Oracles. 
    \item We introduce a PiT inference model that identifies a PiT conditioned on a query $(O, D, T)$, making it possible to reduce the impact of outlier trajectories. 
    \item We propose a Masked Vision Transformer to model global spatial-temporal correlation in the inferred PiT, enabling more effective and efficient travel time estimation.
    \item We report on extensive experiments on two real-world datasets  that offer evidence that the proposed method is capable of outperforming existing methods.
\end{enumerate}

The remainder of the paper is organized as follows. Section~\ref{sec:related_work} covers related work.
Section~\ref{sec:preliminaries} introduces preliminaries and formalizes the problem.  Sections~\ref{sec:modeling-history-trajectories} and~\ref{sec:travel-time-estimation} detail PiT inference and travel time estimation, respectively. Section~\ref{sec:exp} reports on the empirical study, and Section~\ref{sec:con} concludes.

\section{Related Work}
\label{sec:related_work}
\subsection{Travel Time Estimation}

Travel time estimation (TTE) has important applications. In location-based services, TTE can provide users with information on how to plan their trip or on when their packages will arrive. In urban planning, many data analyses, such as flow prediction and congestion prediction, depending on the results of TTE. TTE solutions can be classified into path-based TTE and ODT-Oracles, where the biggest difference is whether a path is given. 

\paragraph{Path-based travel time estimation}
Early studies implement regression techniques~\cite{ide2011trajectory} or decomposition methods~\cite{wang2014travel} to estimate the travel time of trajectories. The accuracies of these methods are limited due to their limited modeling capacities. More recently, deep learning approaches that are capable of modeling complex spatial-temporal correlations in trajectories have been gaining attention. WDR~\cite{wang2018learning}, DeepTTE~\cite{wang2018will}, DeepETA~\cite{wu2019deepeta}, TADNM~\cite{xu2020tadnm}, and CompactETA~\cite{fu2020compacteta} all utilize recurrent neural networks~\cite{hochreiter1997long,DBLP:conf/emnlp/ChoMGBBSB14} to model the sequential spatial-temporal information embedded in trajectories. 
Considering travel time as distributions rather than scalar values, DeepGTT~\cite{li2019learning} implements a variational encoder to model the travel time distributions of trajectories. 
To further improve the prediction performance, TAML~\cite{xu2021taml}, DRTTE~\cite{yang2022multitask}, and WDDRA~\cite{gan2021travel} implement a multitask learning framework to make the estimation of timestamps more accurate. 
DFTTE~\cite{sun2022deep} proposes a fusion network to merge multiple sources of information for prediction. MetaTTE~\cite{wang2022fine} incorporate meta-learning technique to generalize travel time estimation on multi-city scenarios. 
STDGCN~\cite{jin2021hierarchical} employs the neural architecture search techniques to identify the optimal network structure for estimation.

\paragraph{{Path-routing methods}}
{
The superiority of path-based TTE methods is enabled by their affiliated trajectories. But in the scenario of an ODT-Oracle, paths are unknown to the model, rendering path-based methods inapplicable. 
One plausible solution to this issue involves leveraging path-routing methods to infer the potential paths between a given origin-destination pair.
Classical algorithms such as Dijkstra's algorithm~\cite{johnson1973note} and various alternative routing methods~\cite{liu2017finding} primarily focus on determining the paths associated with minimal travel costs, yet their calculated routes may deviate from the actual paths drivers would choose.
DeepST~\cite{li2020spatial} makes use of historical travel behavior derived from trajectory data, thereby enhancing the accuracy of generated paths.
It's noteworthy that route recovery methods~\cite{wu2016probabilistic,chondrogiannis2022history} and map-matching algorithms~\cite{chao2020survey}, though bearing structural similarity to path-routing methods, often demand specific details such as arrival time or comprehensive GPS sequences, which are impractical in the context of the ODT-Oracle.
}

\paragraph{ODT-Oracles}
The motivation for building an ODT-Oracle comes from problems related to path-based travel time estimation. During estimation, only origin and destination locations and departure times are given, and it is challenging to build an accurate ODT-Oracle since the underlying trajectories are unknown. TEMP~\cite{wang2019simple} averages the travel times of historical travels and does not contain any learnable parameter. ST-NN~\cite{jindal2017unified} proposes to learn a non-linear mapping of origin-destination location pairs to the corresponding travel times using neural networks. MURAT~\cite{li2018multi} extends the input features with embeddings from road segments, spatial cells, and temporal slots.
Nevertheless, the underlining trajectory of a trip is highly related to the travel time. All the above methods ignore the correlations between origin-destination location pairs and travel trajectories in historical data, so their prediction accuracies are limited. DeepOD~\cite{yuan2020effective} addresses this problem by incorporating historical trajectories during training and making the embeddings of origin-destination pairs and travel trajectories close in the latent space. However, it is sensitive to outliers in historical trajectories and is unable to provide explainable travel times. 

\subsection{Diffusion Models}
The diffusion model is a generative model recently popularized in computer vision~\cite{song2019generative,ho2020denoising,rombach2022high}. Coming from nonequilibrium thermodynamics~\cite{jarzynski1997equilibrium,sohl2015deep}, diffusion models have a strong theoretical background and have been proven to perform excellently in image generation. There are two Markov processes in diffusion models: the forward diffusion process and the reverse denoising diffusion process. The forward diffusion process adds noise to a clean image through a pre-defined noise schedule until it turns into a Gaussian noise. The reverse denoising diffusion process removes noise from the noisy image step-by-step until the clean image is recovered. After training, one can utilize the reverse process to randomly generate images that follow the distribution of the training dataset. 

Due to the effectiveness and flexibility of diffusion models, many ongoing efforts aim to migrate them into other domains and tasks, such as time series imputation~\cite{tashiro2021csdi}, audio synthesis~\cite{kong2020diffwave}, shape generation~\cite{zhou20213d} and language modeling~\cite{austin2021structured}. In this study, we exploit diffusion models as a powerful means of modeling correlations from historical trajectories. Specifically, we propose a conditioned diffusion model for accurate and explainable ODT-Oracle.

\section{Preliminaries} \label{sec:preliminaries}

\subsection{Definitions} \label{sec:def}

\begin{definition}[Trajectory] 
A trajectory $\mathcal T$ is a sequence of timestamped GPS points: $\mathcal T= \langle (g_1, t_1), (g_2, t_2), \dots, (g_{|\mathcal T|}, t_{|\mathcal T|}) \rangle$, where $g_i=(\mathrm{lng}_i,\mathrm{lat}_i), i=1,\dots,|\mathcal T|$ denotes $i$-th GPS point, and $|\mathcal T|$ denotes the total number of GPS points in the trajectory.
\end{definition}

\begin{definition}[Pixelated Trajectory] 
\label{def:pit}
Given an area of interest on the map, usually, the area covering all historical trajectories, we split the longitude and latitude equally with a total number of $L_G$ segments, resulting in a total of ${L_G}^2$ spatial cells. A Pixelated Trajectory~(PiT), $X\in \mathbb R^{L_G\times L_G\times C}$, is represented as a tensor, where $C$ is the number of channels. Each trajectory has a corresponding PiT.
\end{definition}

In one PiT, the feature $X[x,y,k]$ records the value of $k$-th feature in cell $(x, y)$. 
We utilize three feature channels, i.e., Mask, Time of the day~(ToD), and Time offset. If the GPS points in a trajectory $\mathcal T$ never fall into the cell $(x, y)$, then all three channels of the corresponding cell are set to $-1$. If a GPS point $(g_i, t_i)$ in $\mathcal T$ is the earliest one that falls into the cell $(x, y)$, we calculate the values of three channels as follows. 

\begin{enumerate}[leftmargin=*]
    \item \textit{Mask.} Indicates whether the trajectory contains GPS points located in this cell or not. $X[x, y, 1]=1$ means there are one or more GPS points in the trajectory that are located in cell $(x, y)$.
    \item \textit{ToD.} A normalized value with range $[-1, 1]$ that denotes when the cell is visited. It can be calculated as $X[x, y, 2]=2\times(t_i \% 86400) / 86400 -1$, where $t_i$ denotes the Unix timestamp of $i$-th GPS point in the trajectory $\mathcal T$ and $86400$ is the total number of seconds in 24 hours. 
    \item \textit{Time offset.} Also, a normalized value with range$[-1, 1]$ indicates the visiting order of this cell in the trajectory. It can be calculated as $X[x, y, 3]=2\times (t_i - t_1) / (t_{|\mathcal T|} - t_1) - 1$, where $t_i$ denotes the Unix timestamp of $i$-th GPS point in the trajectory $\mathcal T$. 
\end{enumerate}

\begin{figure}[t]
  \centering
	\includegraphics[width=1.0\linewidth]{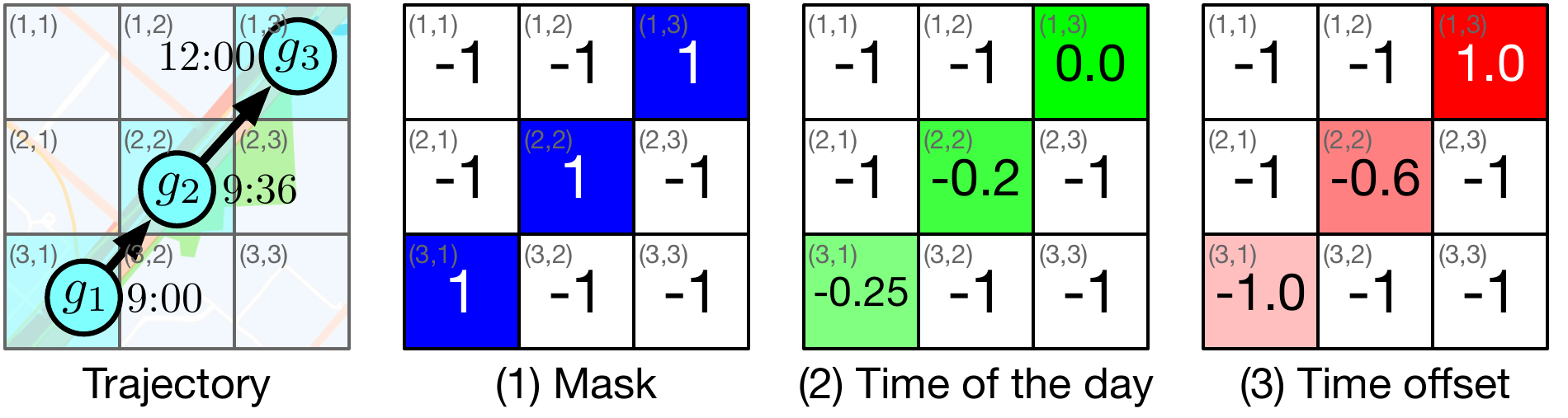}
\caption{Constructing the channels of PiT from GPS points.}
\label{fig:construct-pit}
\end{figure}

\begin{example}
Figure~\ref{fig:construct-pit} demonstrates an example of PiT construction. Suppose we have a trajectory $\mathcal T=\langle (g_1, \text{9:00}), (g_2, \text{9:36}), (g_3, \text{12:00}) \rangle$, and we split the area of interest into $3\times 3$ cells. GPS points $g_1, g_2, g_3$ locate in cell $(3,1), (2,2), (1,3)$, respectively. Thus, we set the mask channels of the three cells as 1. The ToD channels of the cells are calculated as $X[3,1,2]=2\times 9\times 24\times 60/86400-1=-0.25, ~X[2,2,2]=2\times (9\times 24+36)\times 60/86400-1=-0.2, ~X[1,3,2]=2\times (12\times 24\times 60)/86400-1=0.0$, respectively. The time offset channel of the cell $(2,2)$ is calculated as $X[2,2,3]=2\times (9\times24+36) / (12\times24 - 9\times24)=-0.6$, while $X[3,1,3]=-1.0, X[1,3,3]=1.0$. For other cells where no GPS point is located in, all their channels are set to $-1$. 
In this way, we obtain a PiT, $X\in\mathbb R^{3\times 3\times 3}$, corresponding to the trajectory $\mathcal T$.
\end{example}

By representing historical trajectories using PiT rather than using raw sequential features, our method can focus on differences in trajectories that will actually affect the travel time, rather than minor diversities that might cause disturbances. On the other hand, transforming sequential trajectories into equally-sized images is better suited for the two-stage framework we introduce later.

\begin{definition}[ODT-Input]
\label{sec:od-query}
The ODT-Input for an ODT-Oracle, $odt$, is a tuple that consists of three elements: $odt=(g_o, g_d, t_o)$, where $g_o$ and $g_d$ are the GPS coordinates of the origin and destination, respectively. $t_o$ denotes the departure time. 
\end{definition}

\subsection{Problem Formulation}

\noindent
\textbf{OD Travel Time Oracle.} Given the set of historical trajectories $\mathbb{T}$, we aim to learn ODT-Oracle $f^{\mathbb{T}}_\theta$ to estimate the travel time $\Delta t$ and the PiT $X$ for any future ODT-Input, $odt$. To be noted, $odt$ in the query does not require to appear in $\mathbb{T}$. Formally, we have:
\begin{equation}
     odt \xrightarrow{f^{\mathbb{T}}_\theta(\cdot)} \Delta t, X
\end{equation}

\begin{figure*}[t]
  \centering
	\includegraphics[width=0.95\linewidth]{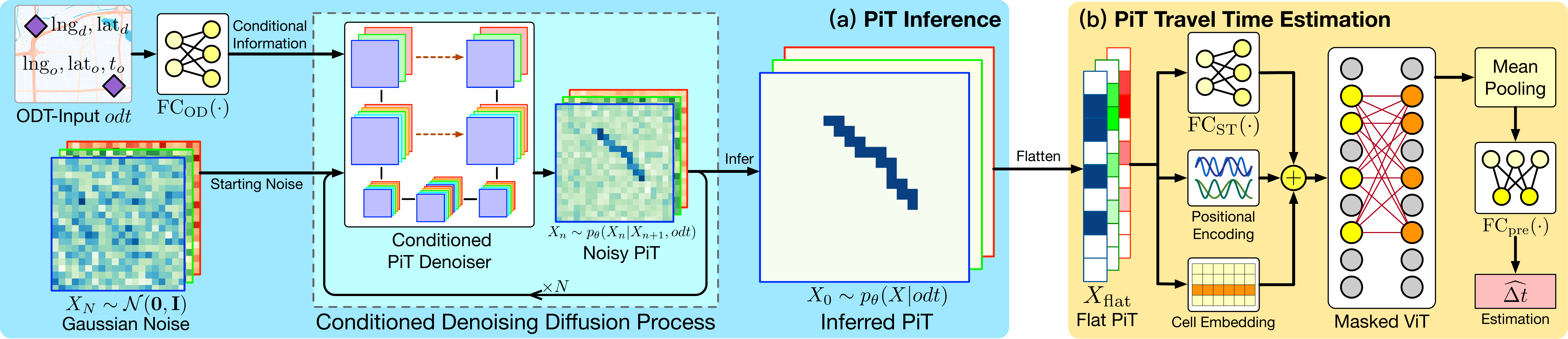}
	\caption{The two-stage framework of DOT.}
	\label{fig:overall-framework}
\end{figure*}

\subsection{Method Overview}
In this work, we propose a \textit{\underline{D}iffusion-based \underline{O}rigin-destination \underline{T}ravel Time Estimation}~(DOT) method to build an accurate and explainable ODT-Oracle, which follows a two-stage framework. 
The first stage is the PiT inference stage, shown in Figure~\ref{fig:overall-framework}(a); the second stage is the PiT travel time estimation stage, shown in Figure~\ref{fig:overall-framework}(b).

In the PiT inference stage, we try to infer the PiT corresponds to a given ODT-Input. The ODT-Input is considered the conditional information incorporated into a conditioned PiT denoiser. 
The denoiser samples from standard Gaussian noise at the beginning. Then, it produces the inferred PiT conditioned on the ODT-Input through a multi-step conditioned denoising diffusion process.

In the PiT travel time estimation stage, we estimate the travel time based on the inferred PiT. The PiT is first flattened and mapped into a feature sequence to capture the global spatial-temporal correlation better. To improve the model's efficiency, we propose a Masked Vision Transformer~(MViT) to estimate the travel time. 

We explain the proposed method in detail in the following sections. The PiT inference stage is introduced in Section~\ref{sec:modeling-history-trajectories}, and the PiT travel time estimation stage is introduced in Section~\ref{sec:travel-time-estimation}.

\begin{figure*}[!t]
  \centering
	\includegraphics[width=0.95\linewidth]{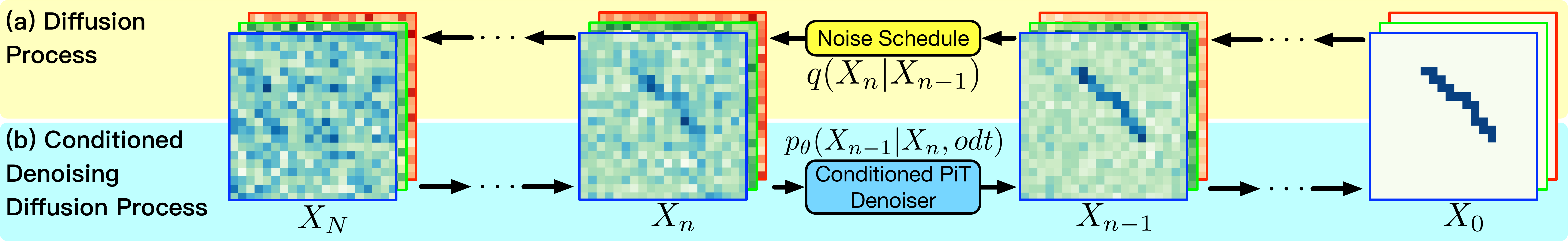}
	\caption{Two Markov processes in the diffusion-based conditioned PiT inference framework.}
	\label{fig:diffusion-model}
\end{figure*}

\section{Modeling Historical Trajectories Through PiT Inference}
\label{sec:modeling-history-trajectories}

\subsection{Diffusion-based PiT Inference}
\label{sec:diffusion-inference-of-trajectory}
Given the origin-destination location pair of a future trip, the travel time is highly related to the route that a driver takes. 
However, the actual route is unavailable at the time of estimating. To achieve accurate ODT-Oracle, we aim to comprehensively learn the correlation between ODT-Inputs and travel trajectories from historical trips. Then, we can utilize the learned correlation to infer travel time based on possible trajectories given the future ODT-Inputs. 

Suppose we have the historical trajectory, denoted as $\mathcal T$, and the corresponding ODT-Input, $odt=(g_o, g_d, t_o)$ of the trip. We aim to learn a posterior probability $p(\mathcal T|odt)$, which represents the relationships between $odt$ and $\mathcal T$ from the set of historical trajectories $\mathbb T$. 
Since we represent a trajectory $\mathcal T$ using the PiT, denoted as $X$, which is stated in Section~\ref{sec:def}, the probability can be written as $p(X|odt)$. However, this probability is unknown in reality, so we aim to learn it through a set of learnable parameters $\theta$. The estimated probability is then denoted as $p_\theta(X|odt)$. 

There is no doubt that the learning model $p_\theta$ is crucial to achieving accurate ODT-Oracle. If the inferred travel trajectory from $p_\theta$ is very different from the expected trajectory given a future ODT-Input, then the estimated travel time cannot be good. 
In this paper, we build our learning model based on Denoising Diffusion Probabilistic Models (DDPM)~\cite{ho2020denoising}. 
Since DDPM models the unconditioned data distribution $p(X)$, the generated data is not restricted, which can be any signal from the underlying data distribution. For example, it can generate various PiTs, whose origins and destinations differ. However, in our case, we aim to model the data distribution conditioned on ODT-Input $odt$. Therefore, we propose a diffusion-based conditioned PiT inference framework, which consists of a diffusion process and a conditioned denoising diffusion process. We detail these two processes in the following Sections. 

\subsubsection{The diffusion process for adding noise to PiTs} 
\label{sec:diffusion-process}
Intuitively, the diffusion process is to add noise to a signal step by step until we reach a simple prior distribution, e.g., Gaussian distribution. Figure~\ref{fig:diffusion-model}(a) gives a clear illustration of how the diffusion process works in our paper, i.e., we try to obtain a noisy PiT, $X_N$, by giving a clear PiT, $X_0$, in a total of $N$ diffusion steps. A single step of this diffusion process can be formulated as follows. 
\begin{equation}
  q(X_n|X_{n-1})=\mathcal N(X_n; \sqrt{1-\beta_n}X_{n-1},\beta_n \mathbf{I}),
  \label{eq:one-step-diffusion-process}
\end{equation}
where $\mathcal N (\mu;\Sigma)$ is the Gaussian distribution with mean value $\mu$ and covariance matrix $\Sigma$, $\beta_n$ is the coefficient used for controlling the noise level in the $n$-th step. In practice, $\beta_n$ is often fixed for every step and follows a monotonic schedule so that the added noise level increases with $n$. We follow the linear schedule used in DDPM~\cite{ho2020denoising}, where $\beta_n$ scales linearly from 0.0001 to 0.02 with $n$.

Starting from the clear PiT $X_0$, we can get the noisy PiT in the $n$-th step by:
\begin{equation}
  q(X_n|X_0)=\prod_{m=1}^{n} q(X_m|X_{m-1})
  \label{eq:multiply-forward-diffusion-process}
\end{equation}

Since the noise level in each step is fixed and the added noises follow Gaussian distribution, we can simplify Equation~\ref{eq:multiply-forward-diffusion-process} into the following form by utilizing the property of Gaussian distribution.
\begin{equation}
  q(X_n|X_0)=\mathcal N(X_n; \sqrt{\overline{\alpha}_n}X_0, (1-\overline{\alpha}_n) \mathbb I),
  \label{eq:simplified-forward-diffusion-process}
\end{equation}
where $\alpha_n=1-\beta_n$ and $\overline{\alpha}_n=\prod_{m=1}^{n} \alpha_m$. Therefore, we can sample the noisy PiT at any step in the diffusion process by adding a specific noise, i.e., $q(X_n|X_0)$, to the original $X_0$, without adding noise step-by-step.

Finally, the noisy PiT in the last step of the diffusion process follows a standard Gaussian noise, which is represented as follows. 
\begin{equation}
  X_N\thicksim q(X_N) = \mathcal N(X_N; \mathbf 0, \mathbf I)
  \label{eq:standard-gaussian-noise}
\end{equation}

Here, $X_N$ is a good start for the following PiT inference process since it contains no prior information. In other words, we can randomly sample noise from the standard Gaussian distribution as the starting point for the inference process.

\subsubsection{The conditioned denoising diffusion process for PiT inference} 
To infer a PiT given a future ODT-Input, we implement a reverse diffusion process under the prior information of ODT-Input. More specifically, we gradually remove noise from the noisy PiT, until we retrieve a noise-free PiT of the travel trajectory corresponding to the ODT-Input. The process is formally called the conditioned denoising diffusion process, which is illustrated in Figure~\ref{fig:diffusion-model}(b). 
A single step of the conditioned denoising diffusion process can be formulated as follows. 
\begin{equation}
\label{eq:one-step-denoising-diffusion-process}
  p(X_{n-1}|X_n, odt) = \mathcal N(X_{n-1}; \boldsymbol \mu_{n-1}, \boldsymbol \Sigma_{n-1}),
\end{equation}
where $\boldsymbol \mu_{n-1}$ and $\boldsymbol \Sigma_{n-1}$ are the mean and variance at the step of $n-1$, and $X_{n-1}$ follows a conditional probability relies on the noisy PiT from the previous step $X_n$ and the ODT-Input $odt$. However, we are unaware of the actual form of $p$ in practice, so we implement two neural network modules to learn the mean and variance. Thus, Equation~\ref{eq:one-step-denoising-diffusion-process} can be reformulated as follows. %
\begin{equation}
  p_\theta(X_{n-1}|X_n, odt) = \mathcal N(X_{n-1}; \boldsymbol \mu_\theta(X_n, n, odt), \boldsymbol \Sigma_\theta(X_n, n, odt)),
  \label{eq:parameterized-denoising-diffusion-process}
\end{equation}
where $\boldsymbol \mu_\theta(\cdot)$ and $\boldsymbol \Sigma_\theta(\cdot)$ denotes the mean and variance estimated by the neural network parameterized by $\theta$. 
The complete PiT inferring process can be formulated as follows. 
\begin{equation}
  p_\theta(X_0|X_N, odt) = \prod_{n=1}^{N} p_\theta(X_{n-1}|X_n, odt)
\end{equation}

As is denoted in Equation~\ref{eq:standard-gaussian-noise}, $X_N$ can be sampled from a standard Gaussian distribution, and $odt$ is the input from the user. Therefore, we can infer a PiT based on these two inputs. We detail the process of PiT inference in Algorithm~\ref{alg:inferring}.

\begin{algorithm}[t]
  \caption{Inferring PiT given ODT-Input}
  \label{alg:inferring}
  \begin{algorithmic}[1]
    \Require A simulated probability $p_\theta(X_{n-1}|X_n,odt)$, a future ODT-Input $odt$
    \State Initialize $n\gets N, X_n\thicksim \mathcal N(\mathbf 0, \mathbf I)$
    \While {n > 0}
      \State Sample $X_{n-1}\thicksim p_\theta(X_{n-1}|X_n,odt)$
      \State Set $X_n\gets X_{n-1}, n\gets n-1$
    \EndWhile
    \State Obtain generation result $X_0$ as the inferred PiT $X$ representative of the travel trajectory of $odt$.
  \end{algorithmic}
\end{algorithm}

Since we utilize a probabilistic model to infer the most plausible PiT given an ODT-Input, we can remove the impact of outliers. 
The remaining question is how to train the set of parameters $\theta$ with the historical trajectories $\mathbb T$, which we discuss in the following section.

\subsubsection{Re-parameterization and training of the PiT inference model.}
To simplify the training of $\theta$, we follow DDPM~\cite{ho2020denoising} to build the conditioned denoising diffusion process by keeping the variance in Equation~\ref{eq:parameterized-denoising-diffusion-process} fixed at any given step, i.e., $\Sigma_\theta(X_n, n, odt)=\sqrt{\beta_n} \mathbf I$. Therefore, only the mean needs to be estimated. 

Then, we re-parameterize the mean $\boldsymbol \mu_\theta(\cdot)$. Instead of directly predicting $\boldsymbol \mu_\theta(\cdot)$, we aim to predict the added noise at the $n$-th step in the diffusion process. We denote the predicted noise as $\epsilon_\theta(X_n, n, odt)$, such that  $\boldsymbol \mu_\theta(\cdot)$ can be re-parameterized as follows. 
\begin{equation}
\label{eq:mean}
  \boldsymbol \mu_\theta(X_n, n, odt) = \frac{1}{\sqrt{\alpha_n}}(X_n - \frac{\beta_n}{\sqrt{1-\overline{\alpha}_n}}\epsilon_\theta(X_n, n, odt))
\end{equation}

Next, we re-parameterize Equation~\ref{eq:parameterized-denoising-diffusion-process}, and substitute $\boldsymbol \mu_\theta(\cdot)$ with Equation~\ref{eq:mean}, which can be formulated as follows. 
\begin{equation}
  \begin{split}
    X_{n-1} &= \boldsymbol \mu_\theta(X_n, n, odt) + \boldsymbol \Sigma_\theta(X_n, n, odt)\epsilon\\
    &= \frac{1}{\sqrt{\alpha_n}}(X_n - \frac{\beta_n}{\sqrt{1-\overline{\alpha}_n}}\epsilon_\theta(X_n, n, odt)) + \sqrt{\beta_n}\epsilon,
  \end{split}
\end{equation}
where $\epsilon \thicksim \mathcal N(\mathbf 0, \mathbf I)$.

To train the set of parameters $\theta$, we must supervise all steps of the conditioned denoising diffusion process. In practice, we can sample $n$ from a uniform distribution, $U(1,N)$, so that all steps can eventually be trained. 
At the $n$-th step, we use the mean squared error to minimize the difference between the ground truth and the predicted noise, and the loss function can be formulated as follows. 
\begin{equation}
  \begin{split}
    L_n&=\parallel \epsilon - \epsilon_\theta(X_n, n, odt) \parallel ^2\\
    &=\parallel \epsilon - \epsilon_\theta(\sqrt{\overline{\alpha}_n}X_0 + \sqrt{1-\overline{\alpha}_n} \epsilon, n, odt) \parallel ^2,
  \end{split}
\end{equation}
where the noisy PiT, $X_n$, comes from the diffusion process and is calculated using the simplified form presented in Equation~\ref{eq:simplified-forward-diffusion-process}. The detailed training algorithm is presented in Algorithm~\ref{alg:training}.

\begin{algorithm}[t]
  \caption{Training PiT inference model}
  \label{alg:training}
  \begin{algorithmic}[1]
    \Require A noise predictor $\epsilon_\theta$ with a set of learnable parameters $\theta$
    \While {model not converged}
    \State Sample $\mathcal T \in \mathbb T$, calculate the ODT-Input $odt$, and the corresponding PiT $X$ as the original image $X_0$
    \State Sample $n\thicksim \mathrm{Uniform}({1, 2, \dots, N})$
    \State Sample $\epsilon \thicksim \mathcal N(\mathbf 0, \mathbf I)$
    \State Calculate $X_n=\sqrt{\overline{\alpha}_n}X_0 + \sqrt{1-\overline{\alpha}_n} \epsilon$
    \State Take a gradient descent step on 
      $\bigtriangledown_\theta \parallel \epsilon - \epsilon_\theta(X_n, n, odt) \parallel ^2$
    \EndWhile
  \end{algorithmic}
\end{algorithm}

\subsection{Conditioned PiT Denoiser} \label{sec:conditioned-pit-denoiser}

\begin{figure}
  \centering
  \subfigure[The overall architecture of the conditioned PiT denoiser.] {
    \begin{minipage}[t]{0.50\linewidth}
			\centering
			\includegraphics[width=1.0\linewidth]{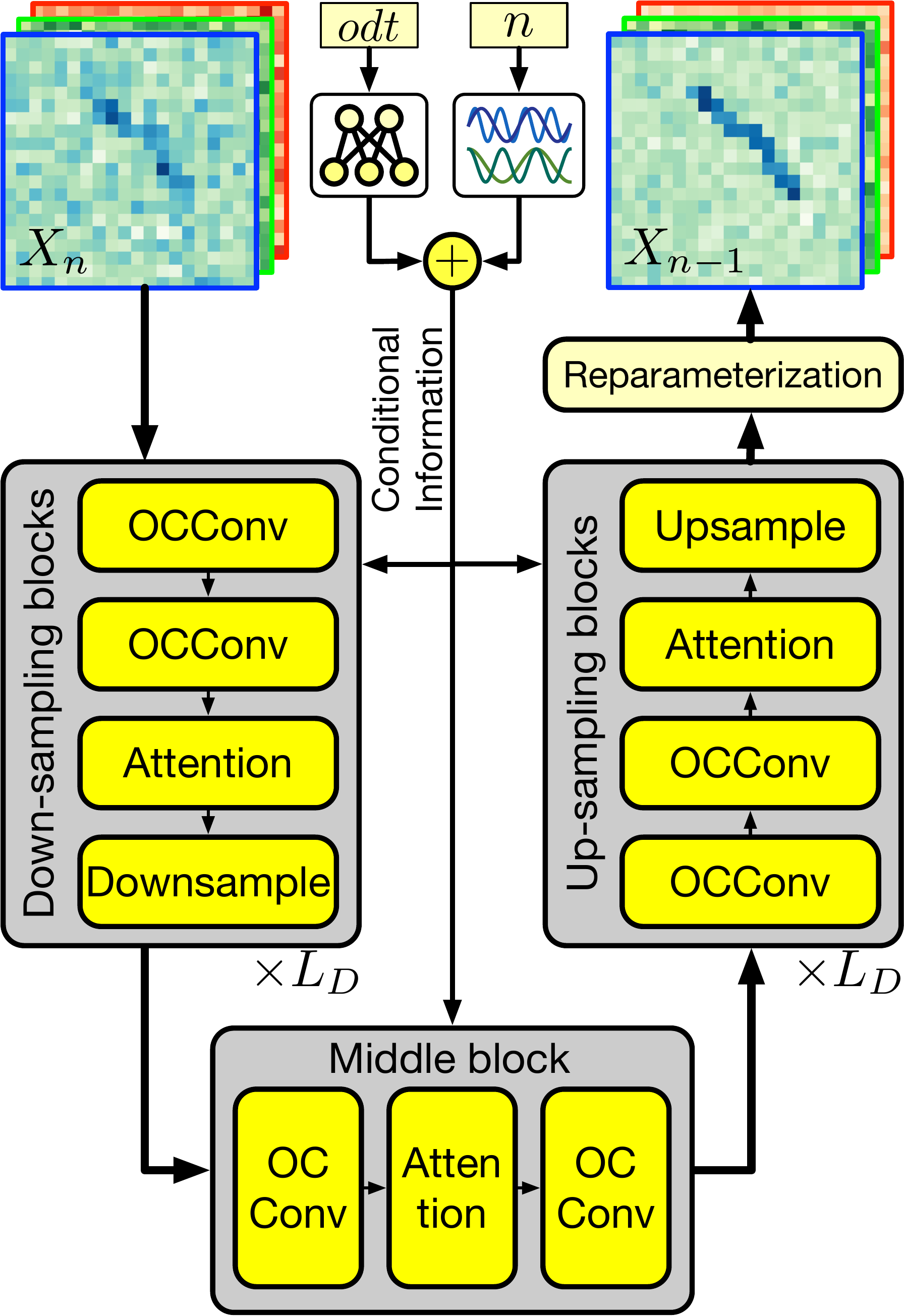}
		\end{minipage}
  \label{fig:pit-denoiser}
  }
  \subfigure[The fuse of conditional information in the OCConv module.] {
    \begin{minipage}[t]{0.42\linewidth}
			\centering
			\includegraphics[width=1.0\linewidth]{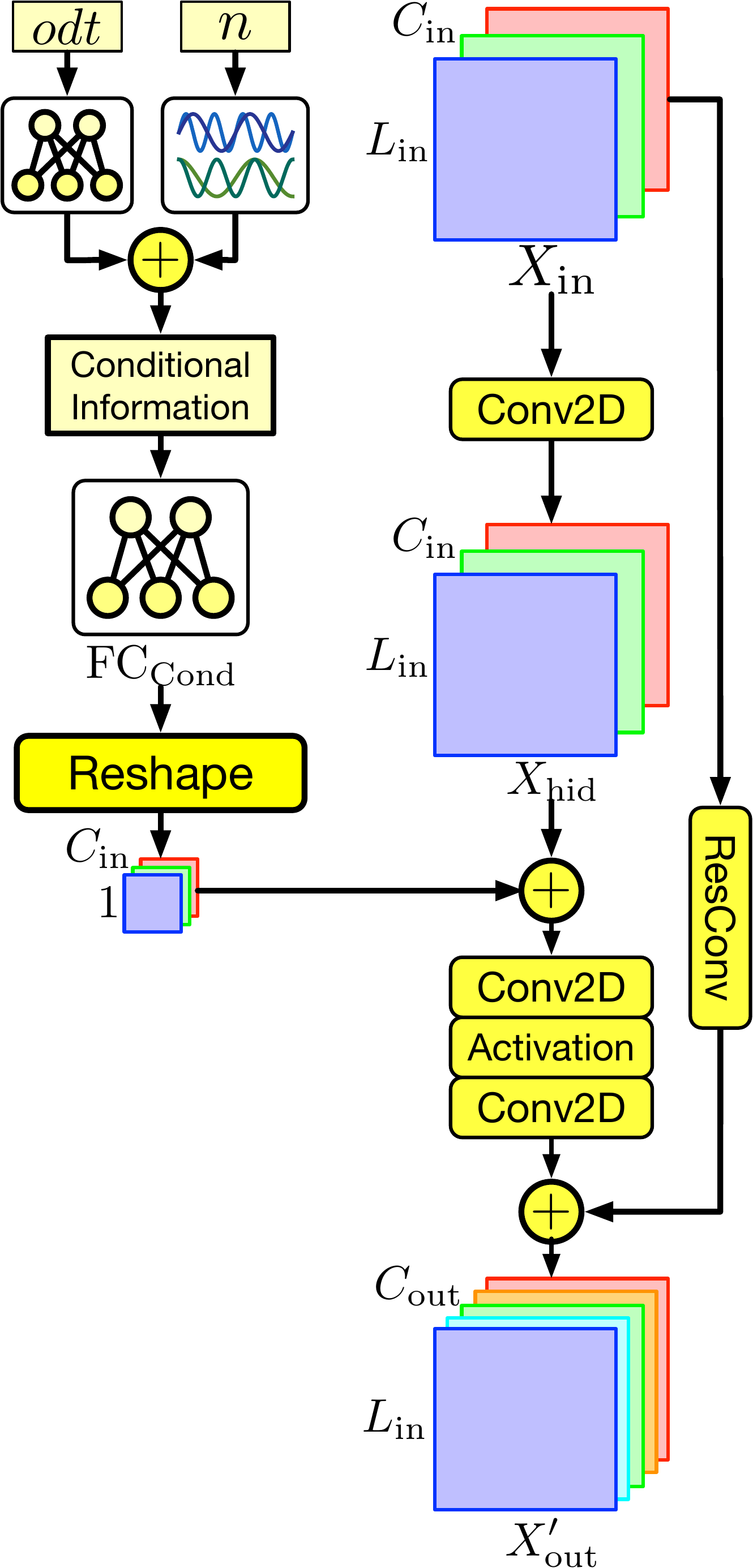}
		\end{minipage}
  \label{fig:condition-fuse}
  }
	\caption{Conditioned PiT denoiser.}
\end{figure}

In this section, in detail, we introduce the implementation of denoiser $\epsilon_\theta(X_n, n, odt)$. Two basic requirements need to be met by the denoiser. The first one is that the predicted noise, i.e., this denoiser's output, should be the same size as the input noisy PiT, $X_n$. The other one is that the denoiser takes input as the noisy PiT $X_n$, the step indicator $n$, and the ODT-Input $odt$. Bearing these requirements, we implement our denoiser based on Unet~\cite{ronneberger2015u}, a neural network widely used in computer vision. Unet is efficient yet powerful due to its bottleneck architecture and residual connection design. Yet, the naive Unet cannot take input as $n$ and $odt$. Thus, we aim to cast these features into latent spaces and fuse them into our Unet-based conditioned PiT denoiser. The overall architecture of our denoiser is shown in Figure~\ref{fig:pit-denoiser}.

We first implement the positional encoding to encode the step indicator $n$ into an embedding vector, $\mathrm{PE}(n)\in\mathbb R^{d}$, which is commonly used in Transformer~\cite{vaswani2017attention}. The detailed encoding operation is formulated as follows: 
\begin{equation}
  \begin{split}
    \mathrm{PE}(n)[2i] &= \sin(n/10000^{2i/d})\\
    \mathrm{PE}(n)[2i-1] &= \cos(n/10000^{2i/d}),\\
  \end{split}
  \label{eq:positional-encoding}
\end{equation}
\noindent
where $d$ is an even value, and $i\in \{1, \dots, d/2\}$ is the dimension indicator of the feature vector. 

We then implement a fully-connected layer to transform $odt$ into a $d$-dimensional latent space, formulated as follows. 
\begin{equation}
  \mathrm{FC}_{\mathrm{OD}}(odt): \mathbb R^5 \rightarrow \mathbb R^d
\end{equation}

Next, we introduce how to fuse these latent representations into the proposed PiT denoiser. Following the Unet~\cite{ronneberger2015u}, our denoiser is formed by $L_D$ layers of down-sampling blocks, one layer of middle block and $L_D$ layers of up-sampling blocks, with residual connections between the down-sampling blocks and the up-sampling blocks. Each down-sampling and up-sampling block contains a sequential stack of two \underline{O}DT-Input \underline{C}onditioned \underline{Conv}olutional (OCConv) modules that are built based on ConvNeXt~\cite{liu2022convnet}, a multi-head dot-product attention module, and a convolutional module for up-sampling or down-sampling. 
The middle block contains two OCConv modules, with an attention module in the middle. 
The down-sampling blocks down-sample the input spatially but expand it channel-wisely, while the up-sampling blocks do the opposite. 
The conditional information $odt$ and $n$ is incorporated into every OCConv module in the denoiser, whose data flow in the OCConv module is illustrated in Figure~\ref{fig:condition-fuse}. 

Specifically, the OCConv module firstly takes input as an image, denoted as $X_{\mathrm{in}}\in \mathbb R^{L_{\mathrm{in}} \times L_{\mathrm{in}} \times C_{\mathrm{in}}}$, which goes through a 2D convolutional layer $\mathrm{Conv2D}$ with all dimensions unchanged, resulting in the hidden state $X_\mathrm{hid}$. 
\begin{equation}
  X_{\mathrm{hid}} = \mathrm{Conv2D}(X_{\mathrm{in}}),~
  X_{\mathrm{hid}}\in \mathbb R^{L_{\mathrm{in}} \times L_{\mathrm{in}} \times C_{\mathrm{in}}}
\end{equation}

Then, we utilize a fully-connected layer to transform the sum of $\mathrm{PE}(n)$ and $\mathrm{FC}_{\mathrm{OD}}(odt)$ into a vector with a size of $C_\mathrm{in}$, which is then added to every pixel in $X_{\mathrm{hid}}$, formulated as follows. 
\begin{equation}
  \begin{split}
    &\mathrm{FC}_{\mathrm{Cond}}(\cdot) : \mathbb R^{d} \rightarrow \mathbb R^{C_\mathrm{in}} \\
    &X_{\mathrm{hid}}'[:, :, i] = X_{\mathrm{hid}}[:, :, i] + \mathrm{FC}_{\mathrm{Cond}}(\mathrm{PE}(n) + \mathrm{FC}_{\mathrm{OD}}(odt))[i],
  \end{split}
\end{equation}
\noindent
where $i = 1, 2, \dots, C_{\mathrm{in}}$. 

Finally, $X_{\mathrm{hid}}'$ goes through a two-layer 2D convolutional network with activation and residual connection to the output state:
\begin{equation}
  \begin{split}
    X_{\mathrm{out}} &= \mathrm{Conv2D}(\sigma(\mathrm{Conv2D}(X_{\mathrm{hid}}')))\\
    X_{\mathrm{out}}' &= X_{\mathrm{out}} + \mathrm{ResConv}(X_{\mathrm{in}}),
  \end{split}
\end{equation}
where $\sigma(\cdot)$ is the activation function, and we use the Gaussian Error Linear Units (GELU)~\cite{hendrycks2016gaussian} in this paper. $\mathrm{ResConv}$ is the residual connection implemented using 2D convolution. The output state $X_{\mathrm{out}}'\in \mathbb R^{L_{\mathrm{in}} \times L_{\mathrm{in}} \times C_{\mathrm{out}}}$ is then fed into the following modules in the denoiser. 
Usually, $C_{\mathrm{out}}=2\cdot C_\mathrm{in}$ for the down-sampling blocks, $C_{\mathrm{out}}= \lfloor C_\mathrm{in}/2 \rfloor$ for the up-sampling blocks.

\section{PiT Travel Time Estimation}
\label{sec:travel-time-estimation}
Once we finish the training of the PiT inference stage detailed in Section~\ref{sec:modeling-history-trajectories}, we aim to propose a PiT travel time estimation module based on the inferred PiT, which is detailed in this Section.

Since the inferred PiT, $X$, is in the pixelated format, it is intuitive to come up with an estimator based on convolutional network networks (CNNs). Yet, CNNs focus on modeling local properties, which is not good at travel time estimation, where capturing the global correlation is essential. Vision Transformer (ViT)~\cite{DBLP:conf/iclr/DosovitskiyB0WZ21} that utilizes self-attention to consider global correlation among all pixels seems to be a good fit in our case. However, many pixels in PiT don't contain valid information, making the vanilla ViT perform poorly in our scenario. Therefore, we propose the Masked Vision Transformer (MViT) to improve the efficiency in both training and estimating significantly.

\subsection{PiT Flatten and Feature Extraction}
Given the inferred PiT $X\in \mathbb R^{L_G\times L_G\times C}$, we firstly flatten it to a sequence with a length of ${L_G}^2$:
\begin{equation}
  \begin{split}
    X_\mathrm{flat} = 
    \langle
      &X[1, 1, :], X[1, 2, :], \dots, X[1, L_G, :], \\
      &X[2, 1, :], X[2, 2, :], \dots, X[2, L_G, :], \\
      &\dots\\
      &X[L_G, 1, :], X[L_G, 2, :], \dots, X[L_G, L_G, :]
    \rangle
  \end{split}
\end{equation}

After flattening, the cell $(x, y)$ in PiT becomes the $(x + (y - 1) * L_G)$-th item in the sequence. $X_{\mathrm{flat}}$ follows the arrangement of pixels rather than chronological order like normal temporal sequences.

We then further extract the spatial-temporal features from $X_\mathrm{flat}$ using three embedding modules:
\begin{enumerate}
  \item \textit{Cell embedding module} $E[\cdot]$. We initialize an embedding matrix $E\in \mathbb R^{{L_G}^2 \times d_E}$, where the column $E[x + (y - 1) * W]$ is the embedding vector of the cell $(x, y)$, that is, the embedding vector of item $X[x, y, :]$ in the flatten sequence. $d_E$ is the embedding dimension. $E$ can be viewed as the innate spatial features of cells.
  \item \textit{Positional encoding module} $\mathrm{PE}(\cdot)$. Since self-attention is order-independent, we encode the position of items in a flattened sequence using the positional encoding introduced in Equation~\ref{eq:positional-encoding}. Here, the encoding vector for item $X[x, y, :]$ is $\mathrm{PE}(x + (y - 1) * W)\in\mathbb R^{d_E}$.
  \item \textit{Latent casting module} $\mathrm{FC}_\mathrm{ST}(\cdot)$. Recall that we design three feature channels for each cell in Section~\ref{sec:def}. We utilize a fully-connected layer to cast them into the latent space: $\mathrm{FC}_{\mathrm{ST}}(X[x, y, :]): \mathbb R^3 \rightarrow \mathbb R^{d_E}$.
\end{enumerate}
The outputs from three modules are summed up to form the latent input vector for each item in $X_\mathrm{flat}$:
\begin{equation}
  \begin{split}
    X_\mathrm{latent}[x, y] = &~E[x + (y - 1) * W] + \\
    &~\mathrm{PE}(x + (y - 1) * W) + \\
    &~\mathrm{FC}_{\mathrm{ST}}(X[x, y, :])
  \end{split}
\end{equation}

Then, the latent sequence, $X_\mathrm{latent}=\langle X_\mathrm{latent}[1, 1], X_\mathrm{latent}[1, 2], \cdots, X_\mathrm{latent}[L_G, L_G] \rangle$ is the input to MViT.

\begin{figure}[t]
  \centering
  \subfigure[Vanilla Vision Transformer] {
    \begin{minipage}[t]{0.47\linewidth}
			\centering
			\includegraphics[width=0.9\linewidth]{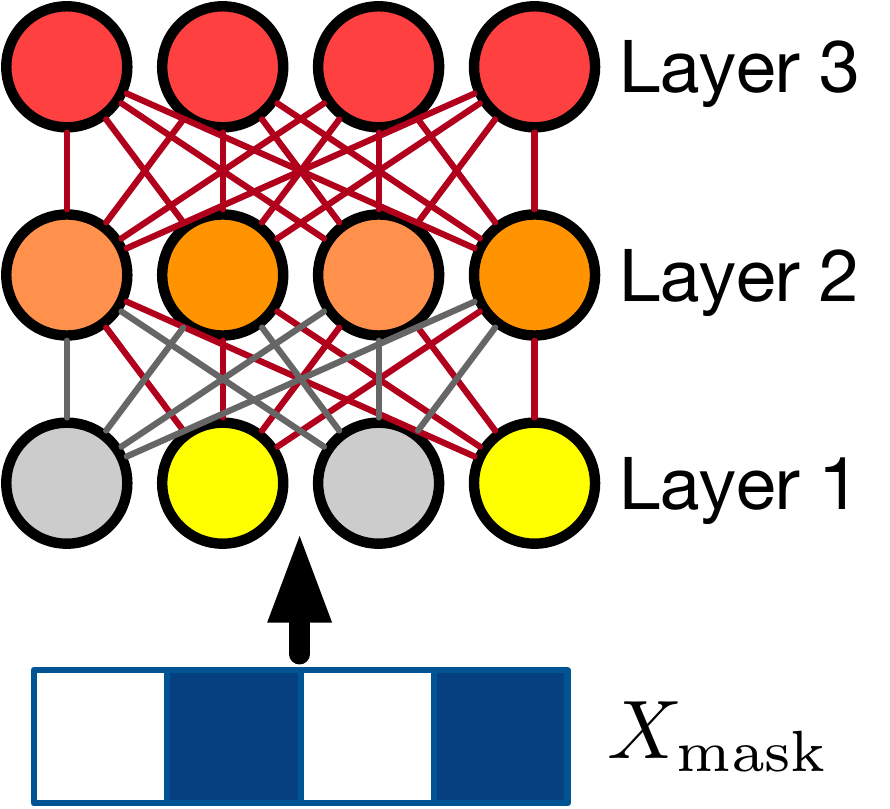}
		\end{minipage}
  \label{fig:vanilla-vit}
  }
  \subfigure[Masked Vision Transformer] {
    \begin{minipage}[t]{0.47\linewidth}
			\centering
			\includegraphics[width=0.9\linewidth]{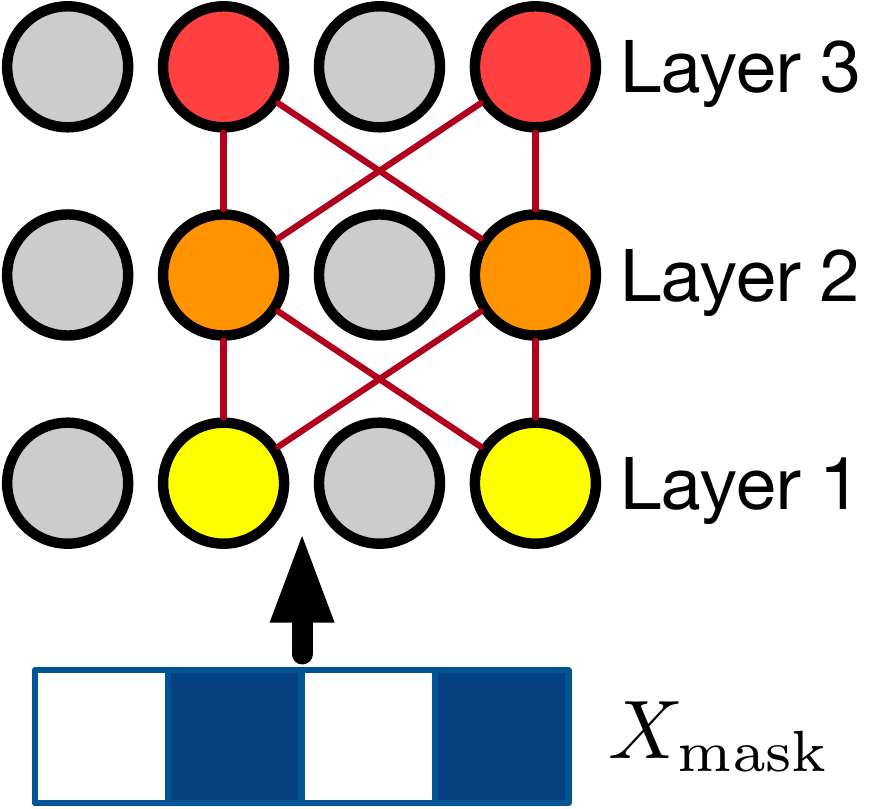}
		\end{minipage}
  \label{fig:masked-vit}
  }
	\caption{Comparison of attention mask scheme between vanilla ViT and MViT.}
	\label{fig:mask-scheme-comparison}
\end{figure}

\subsection{Masked Vision Transformer}
Many items in $X_{\mathrm{latent}}$ don't contain any valid information since their corresponding features in PiT are set to $-1$. In the vanilla vision Transformer (ViT), we can apply an attention mask so that the attention weights of these items are set to zero. Yet, this mask scheme can't improve computational efficiency since the attention weights are still calculated for all items, shown in Figure~\ref{fig:vanilla-vit}. 
This is particularly problematic in our scenario, where PiT covers the full spatial space, but most cells are not visited for one trajectory.

To speed up the calculation on flattened PiT sequences, we aim to implement a more efficient mask scheme. 
We propose the \underline{M}asked \underline{Vi}sion \underline{T}ransformer (MViT). First, we calculate a mask to determine whether one item in $X_\mathrm{latent}$ contains valid information, and we can obtain such a mask through the first channel of PiT:
\begin{equation}
  X_{\mathrm{mask}}[x, y] = \left\{
    \begin{array}{ll}
      \mathrm{True} & \mbox{if $X[x, y, 1] \geqslant 0$} \\
      \mathrm{False} & \mbox{if $X[x, y, 1] < 0$}
    \end{array}
  \right.,
\end{equation}
for $x\in \{1, \dots, L_G\}, y\in \{1, \dots, L_G\}$. $X_\mathrm{latent}[x, y, :]$ contains valid information if $X_\mathrm{mask}[x, y] = \mathrm{True}$. 

Following the Transformer~\cite{vaswani2017attention} and ViT, our MViT is stacked by multiple layers of MViT layer, and each layer contains two modules, a self-attention and a feed-forward network, both with the residual connection. Unlike ViT, self-attention in MViT is only applied to items with valid information. It is equivalent to only retaining the items with valid information to form a masked sequence and only applying self-attention to the masked sequence, as demonstrated in Figure~\ref{fig:masked-vit}. Formally, one MViT layer is calculated as:
\begin{equation}
\begin{split}
    X_\mathrm{out-seq} =& \mathrm{FFN}(\mathrm{Att}(\mathrm{Mask}(X_\mathrm{in-seq}, X_\mathrm{mask})))\\
    \mathrm{Mask}(X_\mathrm{in-seq}, X_\mathrm{mask}) =& \langle X_\mathrm{in-seq}[x, y, :], \\
    &x\in \{1, \dots, L_G\}, y\in \{1, \dots, L_G\}, \\
    &X_\mathrm{mask}[x,y] = \mathrm{True} 
    \rangle,
\end{split}
\end{equation}
where $X_\mathrm{in-seq}, X_\mathrm{out-seq}$ represent the input sequence and output sequence of the MViT layer, $\mathrm{Mask}(X_\mathrm{in-seq}, X_\mathrm{mask})$ denotes the masked input sequence, $\mathrm{Att}(\cdot), \mathrm{FFN}(\cdot)$ denote the multi-head attention and the feed-forward network, respectively. Since we only apply self-attention on items with valid information, the calculation cost depends on the total number of valid items, not the length of the full sequence. Therefore, we can improve the efficiency of flattening PiT sequences significantly.

Then, we stack a total of $L_E$ MViT layers to form the MViT, and calculate the memory sequence using MViT. 
\begin{equation}
  X_\mathrm{latent}' = \mathrm{MViT}(X_\mathrm{latent}, X_\mathrm{mask}),
\end{equation}
where the memory sequence $X_\mathrm{latent}'$ has the same dimension as $X_\mathrm{latent}$, with length equal to the number of valid items in $X_\mathrm{latent}$. 

Next, we apply a mean pooling layer and a fully-connected layer to calculate the result of travel time estimation as follows. 
\begin{equation}
  \widehat{\Delta t} = \mathrm{FC}_\mathrm{pre}(\mathrm{mean}(X_\mathrm{latent}')),
\end{equation}
where the pooling applies to the dimension of sequence length.

Finally, to train the PiT travel time estimation model, we use mean squared error as the loss function to make the prediction result as close to the ground truth as possible, formulated as follows. 
\begin{equation}
  L_{\mathrm{pre}} = \Vert \Delta t - \widehat{\Delta t} \Vert^2
\end{equation}

The two stages, PiT inference and PiT travel time estimation, are trained separately. More specifically, 
after training of the PiT inference model detailed in Algorithm~\ref{alg:training}, the learnable parameters $\theta$ are fixed. During training of the PiT travel time estimation model, $p_\theta(\cdot)$ is only used for inferring, without further parameter update.

By properly utilizing the inferred PiT for travel time estimation, we can achieve high estimation accuracy without the real trajectories that are also not available.

\section{Experiments}
\label{sec:exp}
To evaluate the effectiveness of the proposed DOT framework, we conduct extensive experiments on two real-world trajectory datasets and compare DOT with existing ODT-Oracle methods.

\subsection{Datasets}
In our experiments, we utilize two real-world taxi trajectory datasets collected from the cities of Chengdu from Didi Chuxing\footnote{https://gaia.didichuxing.com/} and Harbin~\cite{li2019learning} in China. We remove trajectories that traveled less than 500 meters or 5 minutes, or more than 1 hour during pre-processing. Then, we filter out sparse trajectories by setting the minimum sampling rate to 80 seconds. The statistics of datasets after pre-processing are listed in Table~\ref{tab:dataset-statistic}.

\begin{table}[t]
    \centering
    \caption{Dataset statistics.}
    \scalebox{0.82}{
    \begin{tabular}{c|ccc}
      \toprule
      Dataset & Chengdu & Harbin \\
      \midrule
      Time span & 11.01--11.10, 2018 & 01.03--01.07, 2015 \\
      Number of trajectories & 1,389,138 & 614,830 \\
      Mean travel time (minutes) & 13.73 & 15.69 \\
      Mean travel distance (meters) & 3,283 & 3,376 \\
      Mean sample interval (seconds) & 29.06 & 44.42 \\
      Area of interest (width$\ast$height km$^2$) & 
        15.32$\ast$15.19 & 18.66$\ast$18.24 \\
      \bottomrule
    \end{tabular}
    }
    \label{tab:dataset-statistic}
\end{table}

\subsection{Comparison Methods}
To prove the superiority of the proposed method, we compare it with twelve baselines. 
Three are routing methods, two are path-based methods, and the others are ODT-Oracle methods.

\subsubsection{Routing Methods}\label{sec:routing-methods}
These methods identify the optimal path on the road network from origin to destination. We provide them with a weighted road network, where the weights represent the average travel time of road segments that is calculated from historical trajectories.
The travel time is the sum of the historical average travel time of the road segments in the identified path.

\begin{itemize}[leftmargin=*]
    \item Dijkstra Shortest Path (\textbf{Dijkstra})~\cite{johnson1973note}: calculates the path between origin and destination with the smallest weights.
    \item Deep Probabilistic Spatial Transition (\textbf{DeepST})~\cite{li2020spatial}: generates the most probable traveling path between origin and destination based on the learned historical travel behaviors.
\end{itemize}

\subsubsection{Path-based Methods}\label{sec:path-based-methods}
These methods predicts the travel time given the travel path. Since the real travel path is absent in the scenario of ODT-Oracle, we feed these methods with the path generated by DeepST.

\begin{itemize}[leftmargin=*]
    \item Wide-Deep-Double Recurrent model with Auxiliary loss (\textbf{WDDRA})~\cite{gan2021travel}: utilizes multi-tasking auxiliary loss to improve the accuracy of travel time estimation.
    \item Automated Spatio-Temporal Dual Graph Convolutional Networks (\textbf{STDGCN})~\cite{jin2021hierarchical}: implement the neural architecture search techniques to automatically identify the optimal network structure.
\end{itemize}

\subsubsection{ODT-Oracle Methods}
These methods aim to predict travel time based on ODT-Input, serving as direct comparison to the proposed method. Four of them are traditional methods, the others are neural network-based methods.

\begin{itemize}[leftmargin=*]
    \item Temporally weighted neighbors (\textbf{TEMP})~\cite{wang2019simple}: averages the travel times of historical trajectories that have a similar origin, destination and departure time.
    \item Linear Regression (\textbf{LR}): learns a linear map from ODT-Inputs to travel times from historical travels.
    \item Gradient Boosted Machine (\textbf{GBM}): a non-linear regression method, which is implemented using XGBoost~\cite{chen2016xgboost}.
    \item Road Network Vertex Embedding (\textbf{RNE})~\cite{huang2021learning}: calculates the shortest path distances between vertices in the embedding space.
    \item Spatial Temporal Neural Network (\textbf{ST-NN})~\cite{jindal2017unified}: jointly predicts the travel distance and time given origin and destination.
    \item MUlti-task Representation learning for Arrival Time estimation (\textbf{MURAT})~\cite{li2018multi}: 
    jointly predicts the travel distance and travel time given origin, destination and departure time.
    \item Effective Travel Time Estimation (\textbf{DeepOD})~\cite{yuan2020effective}: 
    incorporates the correlation between ODT-Inputs and travel trajectories from history through an auxiliary loss during training.
\end{itemize}

\subsection{Experimental Settings} \label{sec:experimental-settings}
For both datasets, we first sort trajectories by their departure date and time. Then, we split them into training, validation and testing sets by 8:1:1. The PiT inference model is trained on the training set for 50 epochs, then used for PiT inference on validation and testing sets. The PiT travel time estimation is trained on the training set, early-stopped on the inferred validation set, and calculated final metrics on the inferred testing set. We use root mean squared error (RMSE), mean absolute error (MAE) and mean absolute percentage error (MAPE) to evaluate the precision of ODT-Oracles.

\begin{table}[t]
    \centering
    \caption{Hyper-parameter range and optimal values.}
    \label{tab:optimal-parameter}
    \begin{threeparttable}
        \begin{tabular}{c|c}
            \toprule
            Parameter & Range \\
            \midrule
            $L_G$ & 10, 15, \underline{20}, 25, 30 \\
            $N$ & 500, \underline{1000}, 1500, 2000 \\
            $L_D$ & 1, 2, \underline{3}, 4 \\
            $d_E$ & 32, 64, \underline{128}, 256 \\
            $L_E$ & 1, \underline{2}, 3, 4 \\
            \bottomrule
        \end{tabular}
        \begin{tablenotes}\footnotesize
            \item[]{\underline{Underline} denotes the optimal value.}
        \end{tablenotes}
    \end{threeparttable}
\end{table}

All methods are implemented using Python and PyTorch~\cite{paszke2019pytorch}. Baselines are implemented following the parameters suggested in their original papers. For the hyper-parameters of the proposed method, we consider 5 key parameters in different modules, their range and optimal values are listed in Table~\ref{tab:optimal-parameter}. It's worth mentioning that all hyper-parameters are chosen based on MAE results on the validation set. We also demonstrate the effectiveness of these hyper-parameters on the testing set later.

During model training, we choose the Adam optimizer and an initial learning rate of 0.001 across the board. We run all experiments on Ubuntu 20.04 servers equipped with Intel(R) Xeon(R) W-2155 CPUs and nVidia(R) TITAN RTX GPUs.

\subsection{Comparison with Baselines}

\begin{table}[t]
    \centering
    \caption{Overall travel time estimation performance of different approaches. }
    \label{tab:overall-result}
    \scalebox{0.94}{
    \begin{threeparttable}
        \begin{tabular}{c|ccc}
            \toprule
            Datasets & \multicolumn{3}{c}{Chengdu/Harbin} \\
            \midrule
            Metric & RMSE (minutes) & MAE (minutes) & MAPE (\%) \\
            \midrule
            Dijkstra
                & {9.677/11.865} & {7.618/8.447} & {48.618/55.261}\\
            {DeepST}
                & {4.717/8.926} & {3.452/5.849} & {27.503/37.772} \\
            \midrule
            {WDDRA}
                & {4.581/8.836} & {3.210/5.705} & {24.553/35.617} \\
            {STDGCN}
                & {4.469/8.679} & {3.104/5.564} & {23.187/\underline{33.771}} \\
            \midrule
            TEMP      & 5.578/10.150 & 4.267/7.891 & 36.611/66.781 \\
            LR        & 6.475/10.290 & 5.036/8.006 & 44.514/67.669 \\
            GBM       & 4.999/9.069  & 3.655/6.748 & 29.636/54.413 \\
            {RNE}
                & {4.624/8.571}  & {3.416/6.245} & {27.660/47.956} \\
            ST-NN     & 3.961/8.492  & 2.803/6.114  & 21.532/45.891 \\
            MURAT     & \underline{3.646}/7.937  & 2.384/5.360 & 18.345/41.128 \\
            DeepOD    & 3.764/\underline{7.859}  & \underline{1.789}/\underline{4.533}  & \underline{14.997}/36.974 \\
            \textbf{DOT (Ours)} & \textbf{3.177}/\textbf{7.462}  & \textbf{1.272}/\textbf{3.213}  & \textbf{11.343}/\textbf{26.698} \\
            \bottomrule
        \end{tabular}
        \begin{tablenotes}\footnotesize
            \item[]{\textbf{Bold} denotes the best result, and \underline{underline} denotes the second-best result.}
        \end{tablenotes}
    \end{threeparttable}
    }
\end{table}

\subsubsection{Overall effectiveness comparison}
Table~\ref{tab:overall-result} demonstrates the comparison between different methods on overall effectiveness. The proposed method consistently shows its performance is superior over the two datasets.

{
The performance of routing methods in estimating travel time is largely dependent on the accuracy of their calculated routes. Dijkstra primarily focuses on finding shortest paths, which results in less accurate travel time estimations. In contrast, DeepST learns travel behavior from historical trajectory data, leading to significantly improved estimation accuracy.
}

{
The travel time estimation accuracy of path-based methods heavily relies on the quality of the input travel paths. 
Both WDDRA and STDGCN exhibit slightly improved performance compared to their path provider, DeepST, as they are capable of learning complex correlations between travel paths and travel time.
STDGCN achieves higher accuracy compared to WDDRA, taking advantage of the neural architecture search technique.
}

The four traditional ODT-Oracle methods, TEMP, LR, GBM, and RNE, tackle the ODT-Oracle problem through feature engineering and kernel design.
These methods face difficulties in modeling the intricate correlations between ODT-Inputs and travel times, and they do not take historical trajectories into account, which is essential for accurate ODT-Oracle predictions.
Among them, the linear method LR performs the worst, suggesting that the spatial-temporal features in ODT-Inputs and travel times do not have a linear correlation.
The history average method TEMP gets the second-worst result, primarily due to the imbalance of historical trips under different ODT-Inputs and the presence of outliers in historical trajectories.
GBM outperforms TEMP and LR, attributable to its relatively higher model capacity.
{RNE achieves the best performance among traditional ODT-Oracle methods, as it incorporates hierarchical embeddings to capture the distances between locations more effectively.}

The four neural network-based ODT-Oracle methods consistently outperform their traditional counterparts.
It indicates that neural network-based methods excel at extracting complex spatial-temporal correlations between ODT-Inputs and travel times, thanks to their higher model capacity. 
Therefore, they can learn more appropriate representations for spatial-temporal features. 
This conclusion is even more obvious when comparing GBM with ST-NN, whose inputs are the same, with only the origin and destination. We can observe that ST-NN has a much better result than GBM. We can also observe that MURAT outperforms ST-NN in both datasets on all metrics. MURAT considers more comprehensive factors, e.g., road network, spatial cells and temporal slots. However, they have not considered taking advantage of the historical trajectories until DeepOD. We can observe that DeepOD can achieve the second best in most cases but is still worse than our method. It indicates that inappropriate handling of outliers in historical trajectories can hurt the performance of the ODT-Oracle.

The proposed method DOT gets the best results on both datasets. We transform raw trajectories into Pixelated Trajectories~(PiTs) to make our model more robust on localized, small differences in trajectories. During training, we explicitly model the spatial-temporal correlation between ODT-Inputs and PiTs from historical trips. During travel time estimation, we infer the most probable PiT given a future ODT-Input and utilize the inferred PiT for accurate travel time estimation. The performance superiority of DOT demonstrates the effectiveness of inferring a robust representative form of travel trajectories and avoiding the impact of outliers in historical trajectories.

\begin{table}[t]
    \centering
    \caption{Scalability of methods on Chengdu, measured by MAPE(\%).}
    \label{tab:scalability}
    \begin{threeparttable}
        \begin{tabular}{c|ccccc}
            \toprule
            Scale & 20\% & 40\% & 60\% & 80\% & 100\% \\
            \midrule
            {Dijkstra} 
                & {57.231} & {54.802} & {53.261} & {52.218} & {48.618} \\
            {DeepST} 
                & {32.635} & {29.700} & {28.864} & {27.848} & {27.503} \\
            \midrule
            {WDDRA}
                & {31.081} & {29.475} & {27.005} & {25.756} & {24.553} \\
            {STDGCN}
                & {30.305} & {28.269} & {26.987} & {25.409} & {23.187} \\
            \midrule
            TEMP   & 56.451 & 49.361 & 46.392 & 41.461 & 36.611 \\
            LR     & 90.412 & 77.206 & 61.451 & 48.652 & 44.514 \\
            GBM    & 43.592 & 38.635 & 34.322 & 32.405 & 29.636 \\
            {RNE}
                & {38.386} & {31.129} & {29.700} & {28.838} & {27.660} \\
            ST-NN  & 27.916 & 24.854 & 23.548 & 22.889 & 21.532 \\
            MURAT  & 24.975 & 22.251 & 20.519 & 19.431 & 18.345 \\
            DeepOD & \underline{18.003} & \underline{17.253} & \underline{16.128} & \underline{15.380} & \underline{14.997} \\
            \textbf{DOT (Ours)}    & \textbf{14.951} & \textbf{14.034} & \textbf{13.014} & \textbf{12.486} & \textbf{11.343} \\
            \bottomrule
        \end{tabular}
        \begin{tablenotes}\footnotesize
            \item[]{\textbf{Bold} denotes the best result, and \underline{underline} denotes the second-best result.}
        \end{tablenotes}
    \end{threeparttable}
\end{table}

\subsubsection{Scalability comparison}
To evaluate the scalability of various methods, we sampled the training set of Chengdu by 20\%, 40\%, 60\%, 80\% and 100\%, then tested the MAPE of different methods trained on the sampled training set, whose results are demonstrated in Table~\ref{tab:scalability}.

Generally speaking, all methods benefit from larger-scale training data. Since larger data have higher density, we can improve the generalization and robustness of the trained models. 
When the scale of training data decreases, the performance downgrade over different methods demonstrates their scalability. The proposed method remains relatively stable and consistently achieves the best result compared with other methods under the same circumstances. We can also observe that our worst performance at scale 20\%, \textit{14.951}, is even better than that of DeepOD at scale 100\%, \textit{14.997}. It indicates that the proposed method can work in more data-scarce scenarios than other methods.

\begin{table}[t]
    \centering
    \caption{Efficiency of methods on Chengdu.}
    \label{tab:efficiency}
    \scalebox{0.93}{
    \begin{tabular}{c|ccc}
        \toprule
        \multirow{2}{*}{Property} & Model size & Training time & Estimation speed \\
        & (Bytes) & (minutes/epoch) & (seconds/K queries) \\
        \midrule
        {Dijkstra}
            & {3.16M} & {-} & {0.95} \\
        {DeepST}
            & {5.40M} & {2.33} & {2.74} \\
        \midrule
        {WDDRA}
            & {6.79M} & {1.43} & {2.42} \\
        {STDGCN}
            & {5.50M} & {2.97} & {3.29} \\
        \midrule
        TEMP   & 4.45M & - & 5.73 \\
        LR     & 0.59K & 0.22 & 0.21 \\
        GBM    & 0.76K & 1.23 & 0.39 \\
        {RNE}
        & {0.78M} & {0.42} & {0.34} \\
        ST-NN  & 0.30M & 0.34 & 0.33 \\
        MURAT  & 7.85M & 1.41 & 1.65 \\
        DeepOD & 6.24M & 1.26 & 1.62 \\
        DOT (Ours) & 7.32M & 3.04/1.22 & 1.85 \\
        \bottomrule
    \end{tabular}
    }
\end{table}

\begin{figure*}[!t]
\setlength{\belowcaptionskip}{0pt}
    \centering
    \subfigure[Model size] {
    \begin{minipage}[t]{0.23\linewidth}
        \centering
        \includegraphics[width=1.0\linewidth]{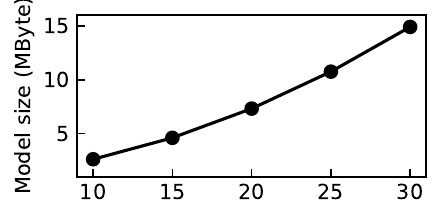}
    \end{minipage}
    \label{fig:grid-model-size}
    }
    \subfigure[Training time of the first stage] {
    \begin{minipage}[t]{0.24\linewidth}
        \centering
        \includegraphics[width=1.0\linewidth]{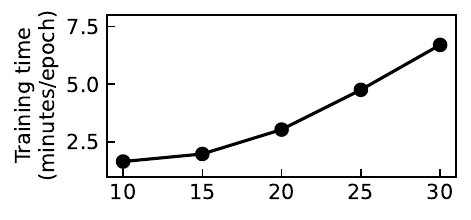}
    \end{minipage}
    \label{fig:grid-train-speed-stage1}
    }
    \subfigure[Training time of the second stage] {
    \begin{minipage}[t]{0.24\linewidth}
        \centering
        \includegraphics[width=1.0\linewidth]{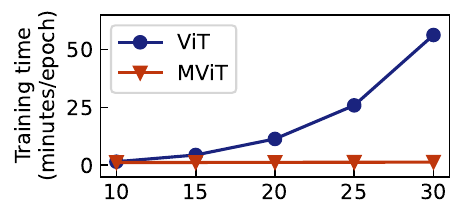}
    \end{minipage}
    \label{fig:grid-train-speed-stage2}
    }
    \subfigure[Estimation speed] {
    \begin{minipage}[t]{0.235\linewidth}
        \centering
        \includegraphics[width=1.0\linewidth]{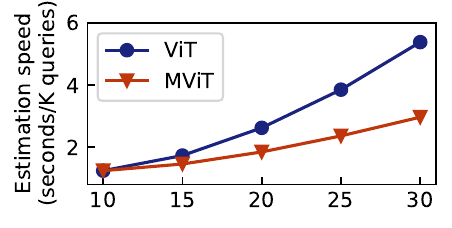}
    \end{minipage}
    \label{fig:grid-estimation-speed}
    }
    \caption{Efficiency impact of grid length $L_G$.}
    \label{fig:hyper-parameters-efficiency}
\end{figure*}

\subsubsection{Efficiency comparison} \label{sec:efficiency-comparison}
To investigate the efficiency, we calculate the model size, training time and estimation speed of different methods on Chengdu. The model size demonstrates the required memory size when running these methods. The training time and estimation speed show the methods' efficiency during training and real-world travel time estimation. Note that the two stages of the proposed method are trained separately, so we give these two training times separately.

{
The storage of the weighted road network in Dijkstra takes up some memory. It doesn't require any training, and expedite techniques to speed up their routing process on road network.
On the other hand, the data-driven routing method DeepST exhibits a relatively slow training and estimation speed, primarily due to the use of RNN sequential model~\cite{hochreiter1997long}. RNNs cannot parallelize on path sequences, which consequently hinders their estimation speed. 
This same reason accounts for the slow estimation speed of the two path-based methods, WDDRA and STDGCN, as they also employ RNNs for processing the input path sequences.
}

Compared with other neural network methods, two traditional methods LR and GBM have smaller model sizes. GBM trains and predicts slower than LR, since it is a non-linear method and contains multiple decision trees for integrated learning. TEMP is an average history method that does not need training. Yet, it needs to cache all historical trips and calculate distances for every pair of trips between queries and training set during prediction. Thus, the model size and prediction speed of TEMP scale nearly linearly with the dataset size, which is problematic for large-scale datasets.

We can observe that ST-NN has the simplest design and smallest model size among neural network-based methods. So, it is the fastest in both training and predicting. 
{RNE stores embeddings for all segments in the road network, resulting in slightly reduced efficiency compared to ST-NN.}
We can also observe that the proposed DOT is relatively slower during its training due to its two-stage framework. However, the prediction speed catches up with the state-of-the-art neural network-based methods since we propose an efficient estimation module, MViT. Since the training phase is always completed offline, we can claim that the efficiency of our method is on par with the existing inference methods.

\subsection{Quantitative Analysis}

\begin{table}[t]
    \centering
    \caption{{Performance of baselines with outlier detection.}}
    \label{tab:baselines-without-outlier}
    \scalebox{0.83}{
    \begin{threeparttable}
        \begin{tabular}{c|ccc}
            \toprule
            Datasets & \multicolumn{3}{c}{Chengdu/Harbin} \\
            \midrule
            Metric & RMSE (minutes) & MAE (minutes) & MAPE (\%) \\
            \midrule
            {Dijkstra+DeepTEA} & 
                {9.641/11.862} & {7.582/8.396} & {48.337/53.949} \\
            {DeepST+DeepTEA} & 
                {4.692/8.901} & {3.416/5.821} & {26.959/37.063} \\
            {WDDRA+DeepTEA} & 
                {4.497/8.584} & {3.140/5.545} & {23.537/34.723} \\
            {STDGCN+DeepTEA} &
                {4.393/8.569} & {3.056/5.501} & {22.812/33.688} \\
            {RNE+DeepTEA} & 
                {4.627/8.403} & {3.447/6.061} & {28.239/45.345} \\
            {ST-NN+DeepTEA} & 
                {3.912/8.427} & {2.740/5.994} & {20.818/43.664} \\
            {MURAT+DeepTEA} & 
                {\underline{3.644}/7.899} & {2.367/5.181} & {17.986/37.728} \\
            {DeepOD+DeepTEA} & 
                {3.763/\underline{7.817}} & {\underline{1.783}/\underline{4.345}} & {\underline{14.835}/\underline{33.127}} \\
            \textbf{DOT (Ours)} & \textbf{3.177}/\textbf{7.462}  & \textbf{1.272}/\textbf{3.213}  & \textbf{11.343}/\textbf{26.698} \\
            \bottomrule
        \end{tabular}
        \begin{tablenotes}\footnotesize
            \item[]{\textbf{Bold} denotes the best result, and \underline{underline} denotes the second-best result.}
        \end{tablenotes}
    \end{threeparttable}
    }
\end{table}

\subsubsection{{Effectiveness of outlier removal.}}
{
We assess the potential performance improvements that can be achieved by combining outlier removal methods with existing baselines. We implement the state-of-the-art outlier detection method DeepTEA~\cite{han2022deeptea} to eliminate outlier trajectories from the training set. Then, we re-train a select set of baselines and evaluate their travel time estimation performance. The results are presented in Table~\ref{tab:baselines-without-outlier}.
}

{
All baselines experience performance improvements, with the exception of RNE on Chengdu. This demonstrates that proper handling of outliers in historical trajectories can be advantageous for achieving accurate travel time estimation. 
Nevertheless, the proposed method continues to outperform these baselines even after outlier removal, for two main reasons.
First, as illustrated in Figure~\ref{fig:pit}, the proposed PiT representation enables our method to more effectively identify outliers that significantly affect travel time estimation, while ignoring negligible differences in trajectories.
Second, we design our PiT inference stage as a diffusion-based generative process, which is more robust at modeling the distribution of historical trajectories with outliers, separating outliers from normal trajectories. This approach has also been demonstrated to be effective in other studies such as~\cite{ho2020denoising,tashiro2021csdi}.
}

\subsubsection{Impact of grid length}
To study the impact of different resolutions of PiT, we vary the grid length, $L_G$, and investigate the efficiency at different grid lengths, i.e., the model size, the training efficiency at the first and second stages, and the inferring efficiency. From Figures~\ref{fig:grid-model-size} and~\ref{fig:grid-train-speed-stage1}, we can observe that the model size and the training time of the first stage increase with the increasing of $L_G$. It is because the PiT becomes larger when $L_G$ increases, bringing increased kernel sizes and filters in the conditioned PiT denoiser. Figure~\ref{fig:grid-train-speed-stage2} shows that the proposed MViT scales well with the increase of $L_G$ compared with the vanilla ViT in training at the second stage. It indicates that the total number of grid cells PiT occupies is almost unchanged, but the proportion of occupied grid cells is becoming less, so MViT can perform much better than ViT. We can see from Figure~\ref{fig:grid-estimation-speed} that MViT also improves the estimation speed significantly compared to ViT.
We can also observe that there is no clear difference between MViT and ViT at $L_G=10$. PiT occupies a much larger proportion of grids when the grid size becomes larger. Yet, the total number of grid cells is the least at $L_G=10$, resulting in the least training time for ViT. 
Finally, we can conclude that the proposed MViT can improve the framework's efficiency in both training and estimating.

\begin{figure}[t]
\centering

\subfigure[Effectiveness of grid length $L_G$] {
    \begin{minipage}[t]{1.0\linewidth}
        \centering
        \includegraphics[width=0.47\linewidth]{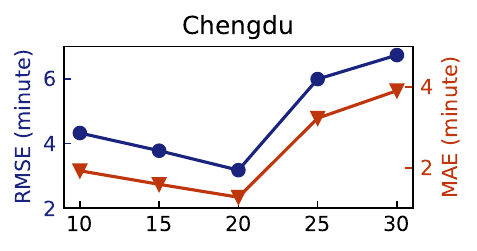}
        \includegraphics[width=0.49\linewidth]{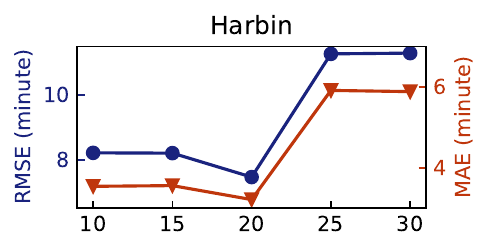}
    \end{minipage}
    \label{fig:parameter-grid-length}
}
\subfigure[Effectiveness of number of diffusion steps $N$] {
    \begin{minipage}[t]{1.0\linewidth}
        \centering
        \includegraphics[width=0.47\linewidth]{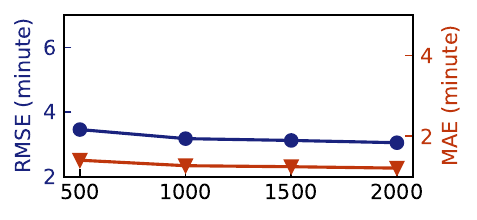}
        \includegraphics[width=0.49\linewidth]{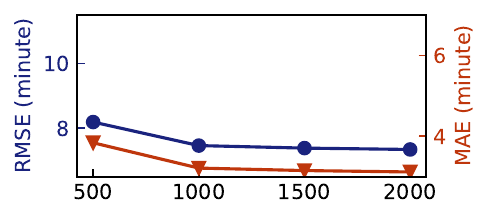}
    \end{minipage}
    \label{fig:parameter-diffusion-step}
}
\subfigure[Effectiveness of number of Unet layer $L_D$] {
    \begin{minipage}[t]{1.0\linewidth}
        \centering
        \includegraphics[width=0.47\linewidth]{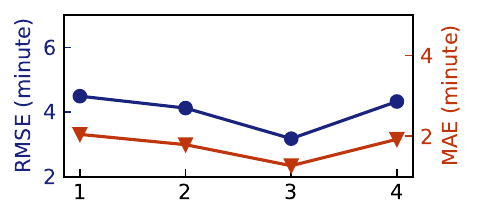}
        \includegraphics[width=0.49\linewidth]{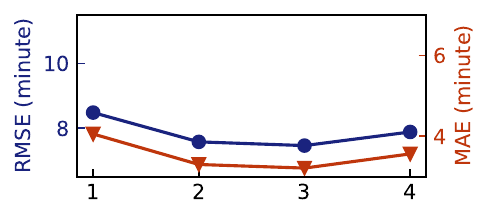}
    \end{minipage}
    \label{fig:parameter-unet-layer}
}
\subfigure[Effectiveness of Transformer embedding dimension $d_E$] {
    \begin{minipage}[t]{1.0\linewidth}
        \centering
        \includegraphics[width=0.47\linewidth]{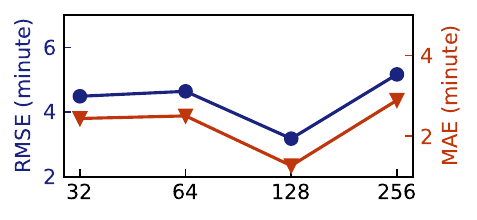}
        \includegraphics[width=0.49\linewidth]{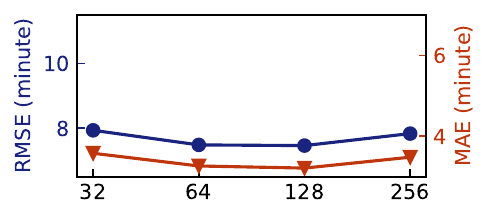}
    \end{minipage}
    \label{fig:parameter-transformer-dmodel}
}
\subfigure[Effectiveness of number of Transformer layers $L_E$] {
\begin{minipage}[t]{1.0\linewidth}
        \centering
        \includegraphics[width=0.47\linewidth]{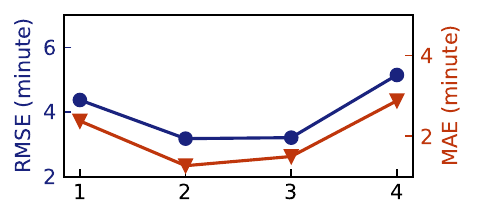}
        \includegraphics[width=0.49\linewidth]{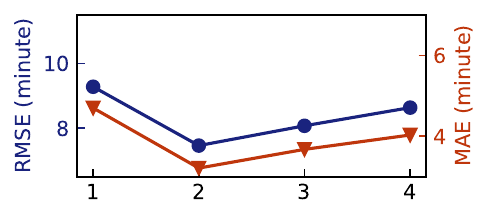}
    \end{minipage}
    \label{fig:parameter-transformer-layer}
}
\caption{Effects of hyper-parameters on ODT-Oracle performance.}
\label{fig:hyper-parameters}
\end{figure}

\subsubsection{Impact of hyper-parameters}
The optimal hyper-parameters listed in Table~\ref{tab:optimal-parameter} are selected with parameter experiments on validation sets. In this section, we further demonstrate the effectiveness of these key hyper-parameters on testing sets, which is shown in Figure~\ref{fig:hyper-parameters}. We have the following observations. 

\begin{enumerate}[leftmargin=*]
    \item As is indicated in Figure~\ref{fig:parameter-grid-length}, it is optimal to set grid length $L_G$ to 20. A smaller grid length results in coarse-grained PiTs, which is insufficient to estimate travel time accurately. On the other hand, a bigger grid length increases the sparsity of PiTs, resulting in the model being more sensitive to negligible differences in trajectories. Interestingly, increasing the grid length hurts the prediction performance more than decreasing it. This shows that it is better to model trajectories into PiTs, which can reduce the disturbance to accurate travel time estimation. 
    \item Figure~\ref{fig:parameter-diffusion-step} indicates that more diffusion steps lead to better results, which is intuitive that we can learn the noise better with more steps. However, the gain becomes less when $N$ is larger than 1000. Therefore, we select $N=1000$ as a good trade-off between effectiveness and efficiency.
    \item $L_D, d_E, L_E$ determine the representation capacity of the PiT inference model and the PiT travel time estimation.
    Figures~\ref{fig:parameter-unet-layer}, \ref{fig:parameter-transformer-dmodel} and~\ref{fig:parameter-transformer-layer} demonstrate their effectiveness respectively. All of them have optimal values. A too-small model struggles to extract the complex spatial-temporal correlations in datasets, while a too-big model leads to overfitting.
\end{enumerate}

\begin{table}[t]
    \centering
    \caption{Effects of features and modules on ODT-Oracle performance.}
    \label{tab:ablation-study}
    \scalebox{0.90}{
    \begin{threeparttable}
        \begin{tabular}{c|ccc}
        \toprule
        Datasets & \multicolumn{3}{c}{Chengdu/Harbin} \\
        \midrule
        Metric & RMSE (minutes) & MAE (minutes) & MAPE (\%) \\
        \midrule
        {Dijkstra+Est.} 
            & {9.182/11.869} & {6.871/8.246} & {41.462/50.488} \\
        {DeepST+Est.} 
            & {4.587/8.879} & {3.170/5.689} & {23.437/33.769} \\
        \midrule
        {Infer.+WDDRA}
            & {3.773/7.958} & {1.801/4.171} & {18.937/31.514} \\
        {Infer.+STDGCN}
            & {3.476/7.611} & {1.664/3.818} & {17.653/29.756} \\
        \midrule
        No-t & 4.325/8.798 & 1.926/4.345 & 16.820/35.973 \\
        No-od & 7.355/10.947 & 4.564/6.333 & 38.879/51.699 \\
        No-odt & 8.466/11.172 & 5.880/6.562 & 49.830/53.331 \\
        No-CE & 3.778/8.584 & 1.591/4.144 & 14.034/34.441 \\
        No-ST & 7.784/11.023 & 5.036/6.427 & 42.850/52.442 \\
        Est-CNN & 6.297/10.389 & 3.500/5.765 & 30.004/47.166 \\
        Est-ViT & \underline{3.229}/\textbf{7.390} & \underline{1.293}/\textbf{3.187} & \underline{11.547}/\textbf{26.484} \\
        \textbf{DOT} & \textbf{3.177}/\underline{7.462} & \textbf{1.272}/\underline{3.213} & \textbf{11.343}/\underline{26.698} \\
        \bottomrule
        \end{tabular}
        \begin{tablenotes}\footnotesize
            \item[]{\textbf{Bold} denotes the best result, and \underline{underline} denotes the second-best result.}
        \end{tablenotes}
    \end{threeparttable}
    }
\end{table}

\subsubsection{Ablation study}\label{sec:ablation-study}
To verify the effectiveness of features and modules in the proposed method, we conduct an ablation study in the following variants of DOT methods and input features. 

\begin{enumerate}[leftmargin=*]
    \item {\textit{Routing+Est.}: combine the routing methods listed in Section~\ref{sec:routing-methods} with the PiT travel time estimation stage of our method.}
    \item {\textit{Infer.+Path-based}: integrates the PiT inference stage with the path-based approaches presented in Section~\ref{sec:path-based-methods}.}
    \item \textit{No-t}: remove the departure time $t_o$ from the ODT-Input $odt$.
    \item \textit{No-od}: remove the origin and destination coordinates from $odt$.
    \item \textit{No-odt}: remove the conditional information $odt$ completely.
    \item \textit{No-CE}: remove the cell embedding module from the PiT travel time estimation.
    \item \textit{No-ST}: remove the latent casting module from the PiT travel time estimation.
    \item \textit{Est-CNN}: replace MViT with a CNN-based estimator.
    \item \textit{Est-ViT}: replace MViT with the vanilla vision Transformer.
\end{enumerate}

We compare the performance of these variants with the DOT method, and the experimental results are given in Table~\ref{tab:ablation-study}. We have the following observations:
\begin{enumerate}[leftmargin=*]
    \item {Combining the routing methods with the proposed PiT travel time estimation stage failed to yield satisfactory results. This is mainly because the routes inferred by these methods are not accurate enough, which is also demonstrated in Table~\ref{tab:inference-compare}.
    Additionally, these routings methods do not generate temporal features, i.e., the second and third channels in PiTs, and these features are instead populated based on historical average travel times between cells. In contrast, the proposed method infers both spatial and temporal features of PiTs based on the learned spatio-temporal information from historical trajectories, which benefits travel time estimation accuracy.}
    \item {Integrating the proposed PiT inference stage with the path-based methods enhances their estimation performance compared to the results in Table~\ref{tab:overall-result}. 
    The improvements can be attributed to the higher accuracy of the inferred routes compared to those produced by DeepST.
    However, the results still do not surpass those obtained by DOT, as the RNN sequential models employed in WDDRA and STDGCN are not as effective as the proposed MViT in modeling spatio-temporal correlations in PiTs, as also shown by other studies of time series mining~\cite{vaswani2017attention,zhou2021informer}.}
    \item Removing features from the conditional ODT-Input reduces the estimation performance, since the inferred PiT cannot accurately correspond to the given ODT-Input.
    \item Removing the embedding modules from the PiT travel time estimation also makes prediction worse, since the spatial-temporal information embedded in PiT is not comprehensively utilized.
    \item Comparing results from Est-CNN and DOT proves that our Transformer-based MViT is more effective than CNN-based methods when dealing with PiT. Comparing results from Est-ViT and DOT shows that the estimation performance of MViT is very close to ViT. Combining the efficiency comparison in Figure~\ref{fig:hyper-parameters-efficiency}, the proposed MViT can improve the efficiency over ViT while not compromising on estimation performance.
\end{enumerate}

\subsection{Analysis on the Explainability}
To evaluate the explainability of the proposed method, we first conduct some quantitative experiments to demonstrate the accuracy of the inferred trajectories. We then conduct a case study to visualize the inferred trajectories.

\begin{table}[t]
    \centering
    \caption{PiT inference accuracy of our model.}
    \label{tab:inference-accuracy}
    \begin{tabular}{c|cc}
        \toprule
        Datasets & \multicolumn{2}{c}{Chengdu/Harbin} \\
        \midrule
        Metric & RMSE & MAE \\
        \midrule
        Overall            & 0.196/0.181 & 0.027/0.023 \\
        Channel 1 (Mask)   & 0.271/0.224 & 0.039/0.028 \\
        Channel 2 (ToD)    & 0.128/0.183 & 0.016/0.024 \\
        Channel 3 (Offset) & 0.159/0.123 & 0.025/0.016 \\
        \bottomrule
    \end{tabular}
\end{table}

\begin{table}[t]
    \centering
    \caption{Accuracy of route inferences.}
    \label{tab:inference-compare}
    \scalebox{0.95}{
        \begin{tabular}{c|ccc}
            \toprule
            Datasets & \multicolumn{3}{c}{Chengdu/Harbin} \\
            \midrule
            Metric & Pre(\%) & Rec(\%) & F1(\%) \\
            \midrule
            {Dijkstra}
                & {\underline{68.918}/45.459} & {31.310/42.525} & {42.065/39.993} \\
            {DeepST}
                & {59.755/\underline{74.519}} & {\underline{55.776}/\underline{62.907}} & {\underline{56.911}/\underline{66.029}} \\
            \textbf{DOT (Ours)} & \textbf{87.890}/\textbf{88.190} & \textbf{88.684}/\textbf{88.982} & \textbf{88.280}/\textbf{88.584} \\
            \bottomrule
        \end{tabular}
    }
\end{table}

\begin{figure}[t]
\setlength{\belowcaptionskip}{0pt}
  \centering
	\includegraphics[width=0.8\linewidth]{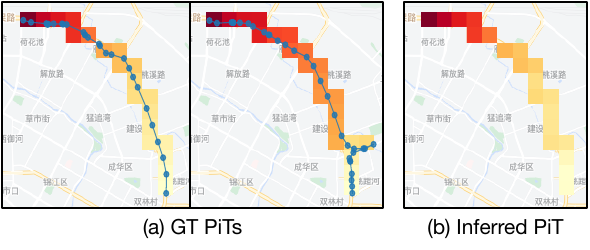}
	\caption{Trajectories for the same OD pair and departure at the same time of the day.}
	\label{fig:case-slight-difference}
\end{figure}

\begin{figure}[t]
\setlength{\belowcaptionskip}{0pt}
  \centering
	\includegraphics[width=1.0\linewidth]{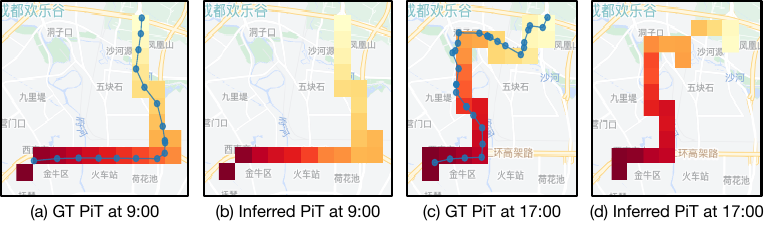}
	\caption{Trajectories for the same OD pair that depart at different times of the day.}
	\label{fig:case-different-departure-time}
\end{figure}

\subsubsection{PiT inference accuracy}\label{sec:pit-inference-accuracy}
We conduct experiments on both datasets to verify the accuracy of the inferred PiTs. Specifically, given the ODT-Inputs from the testing set, 
we calculate the RMSE and MAE between the inferred PiTs and the ground truth ones, whose results are listed in Table~\ref{tab:inference-accuracy}.

{We also compare the accuracy of the inferred routes from DOT with the planned routes in Dijkstra and DeepST. Specifically, the routes planned by these methods are transformed into the same form as the first (mask) channel in PiT.}
We then compare the accuracy of the mask channels calculated by these methods with those inferred by our model, using precision (Pre), recall (Rec), and F1 scores as metrics. The results are presented in Table~\ref{tab:inference-compare}.

We can observe that we achieve a pretty accurate PiT inference and route inference on both datasets, which helps us to achieve good performance in the second stage of travel time estimation. 
Accurate route inference also means that our model can provide the users with the most probable route that is chosen by most drivers, given a future travel plan.

\begin{figure}
  \centering
    \includegraphics[width=1.0\linewidth]{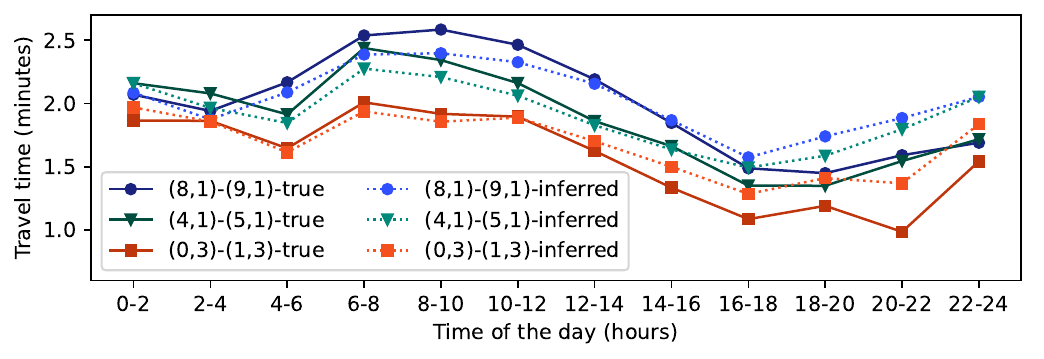}
    \caption{{Average travel time between spatial cells during a day.}}
    \label{fig:avg-time-between-cells}
\end{figure}

\subsubsection{Case study}\label{sec:case-study}
We conduct two case studies on the testing set of Chengdu. 

For the first case study, we visualize the raw trajectories and the third channel~(time offset) of both the ground truth and inferred PiTs under certain circumstances. We investigate the following circumstances with trajectories that travel from the same origin to the same destination. 

\begin{enumerate}[leftmargin=*]
    \item \textit{Trajectories depature during the same time}: In Figure~\ref{fig:case-slight-difference}(a), we can observe that two PiTs are almost the same when departing during the same time of the day, where the second PiT has extra cells compared to the first one, which is an outlier. Figure~\ref{fig:case-slight-difference}(b) shows the PiT inferred from the first stage by giving the ODT-Input with the same OD and T. We can observe that the inferred PiT matches the ground truth well, where the small portion of outliers in the second PiT is removed.
    \item \textit{Trajectories departure during different time}: Figure~\ref{fig:case-different-departure-time} demonstrates the case with the same origin-destination pair but a different departure time. It indicates that different travel trajectories can be chosen during different times of the day, which makes a big difference in the travel time estimation. Since the proposed method considers the departure time, it can infer PiTs at different departure times.
\end{enumerate}

{
In the second case study, we examine the evolution of the average travel time between pairs of spatial cells during a day. We select the top-3 most frequently traveled pairs of cells and calculate the average travel time between them by dividing a day into two-hour intervals.
Figure~\ref{fig:avg-time-between-cells} display the average travel time calculated from ground truth trajectories and the inferred PiTs of DOT, where the cells are denoted by their coordinates specified in Defintion~\ref{def:pit}. The travel time calculated from the inferred PiTs varies throughout the day, indicating that our method accounts for the evolving traffic conditions. Furthermore, it closely matches the travel time calculated from ground truth trajectories, signifying the accuracy of the temporal features in the inferred PiTs.
}

In conclusion, the proposed method's conditioned PiT inference model can learn from history and infer PiT given a future ODT-Input. It ensures the accurate performance of travel time estimation, which is highly dependent on the spatial-temporal information of trajectories. In addition, the inferred PiT gives an intuitive overview of the future trip, improving the explainability of the model, and providing users with useful information.

\section{Conclusions}
\label{sec:con}
To build an accurate and explainable ODT-Oracle, we propose a two-stage framework called DOT. 
In the first stage, we propose a conditioned PiT denoiser to implement a PiT inference model. This model can learn from historical trajectories and can infer a PiT given an OD pair and a departure time, which reduce the impact of outlier historical trajectories. 
In the second stage, we propose an MViT to model the global correlation of the inferred PiT efficiently, which can give an accurate estimation of travel time. Comprehensive experiments are conducted on two real-world datasets to demonstrate the performance superiority of the proposed method.

\balance
\bibliographystyle{ACM-Reference-Format}
\bibliography{reference_cr}


\begin{thebibliography}{61}


\ifx \showCODEN    \undefined \def \showCODEN     #1{\unskip}     \fi
\ifx \showDOI      \undefined \def \showDOI       #1{#1}\fi
\ifx \showISBNx    \undefined \def \showISBNx     #1{\unskip}     \fi
\ifx \showISBNxiii \undefined \def \showISBNxiii  #1{\unskip}     \fi
\ifx \showISSN     \undefined \def \showISSN      #1{\unskip}     \fi
\ifx \showLCCN     \undefined \def \showLCCN      #1{\unskip}     \fi
\ifx \shownote     \undefined \def \shownote      #1{#1}          \fi
\ifx \showarticletitle \undefined \def \showarticletitle #1{#1}   \fi
\ifx \showURL      \undefined \def \showURL       {\relax}        \fi
\providecommand\bibfield[2]{#2}
\providecommand\bibinfo[2]{#2}
\providecommand\natexlab[1]{#1}
\providecommand\showeprint[2][]{arXiv:#2}

\bibitem[Austin et~al\mbox{.}(2021)]%
        {austin2021structured}
\bibfield{author}{\bibinfo{person}{Jacob Austin}, \bibinfo{person}{Daniel~D
  Johnson}, \bibinfo{person}{Jonathan Ho}, \bibinfo{person}{Daniel Tarlow},
  {and} \bibinfo{person}{Rianne van~den Berg}.}
  \bibinfo{year}{2021}\natexlab{}.
\newblock \showarticletitle{Structured denoising diffusion models in discrete
  state-spaces}. In \bibinfo{booktitle}{\emph{NeurIPS}},
  Vol.~\bibinfo{volume}{34}. \bibinfo{pages}{17981--17993}.
\newblock


\bibitem[Cai et~al\mbox{.}(2013)]%
        {cai2013fresh}
\bibfield{author}{\bibinfo{person}{Xiaoqiang Cai}, \bibinfo{person}{Jian Chen},
  \bibinfo{person}{Yongbo Xiao}, \bibinfo{person}{Xiaolin Xu}, {and}
  \bibinfo{person}{Gang Yu}.} \bibinfo{year}{2013}\natexlab{}.
\newblock \showarticletitle{Fresh-product supply chain management with
  logistics outsourcing}.
\newblock \bibinfo{journal}{\emph{Omega}} \bibinfo{volume}{41},
  \bibinfo{number}{4} (\bibinfo{year}{2013}), \bibinfo{pages}{752--765}.
\newblock


\bibitem[Chao et~al\mbox{.}(2020)]%
        {chao2020survey}
\bibfield{author}{\bibinfo{person}{Pingfu Chao}, \bibinfo{person}{Yehong Xu},
  \bibinfo{person}{Wen Hua}, {and} \bibinfo{person}{Xiaofang Zhou}.}
  \bibinfo{year}{2020}\natexlab{}.
\newblock \showarticletitle{A survey on map-matching algorithms}. In
  \bibinfo{booktitle}{\emph{ADC}}. Springer, \bibinfo{pages}{121--133}.
\newblock


\bibitem[Chen and Guestrin(2016)]%
        {chen2016xgboost}
\bibfield{author}{\bibinfo{person}{Tianqi Chen} {and} \bibinfo{person}{Carlos
  Guestrin}.} \bibinfo{year}{2016}\natexlab{}.
\newblock \showarticletitle{Xgboost: A scalable tree boosting system}. In
  \bibinfo{booktitle}{\emph{ACM SIGKDD}}. \bibinfo{pages}{785--794}.
\newblock


\bibitem[Cho et~al\mbox{.}(2014)]%
        {DBLP:conf/emnlp/ChoMGBBSB14}
\bibfield{author}{\bibinfo{person}{Kyunghyun Cho}, \bibinfo{person}{Bart van
  Merrienboer}, \bibinfo{person}{{\c{C}}aglar G{\"{u}}l{\c{c}}ehre},
  \bibinfo{person}{Dzmitry Bahdanau}, \bibinfo{person}{Fethi Bougares},
  \bibinfo{person}{Holger Schwenk}, {and} \bibinfo{person}{Yoshua Bengio}.}
  \bibinfo{year}{2014}\natexlab{}.
\newblock \showarticletitle{Learning Phrase Representations using {RNN}
  Encoder-Decoder for Statistical Machine Translation}. In
  \bibinfo{booktitle}{\emph{EMNLP}}. \bibinfo{pages}{1724--1734}.
\newblock


\bibitem[Chondrogiannis et~al\mbox{.}(2022)]%
        {chondrogiannis2022history}
\bibfield{author}{\bibinfo{person}{Theodoros Chondrogiannis},
  \bibinfo{person}{Johann Bornholdt}, \bibinfo{person}{Panagiotis Bouros},
  {and} \bibinfo{person}{Michael Grossniklaus}.}
  \bibinfo{year}{2022}\natexlab{}.
\newblock \showarticletitle{History oblivious route recovery on road networks}.
  In \bibinfo{booktitle}{\emph{SIGSPATIAL}}. \bibinfo{pages}{1--10}.
\newblock


\bibitem[Crainic and Laporte(1997)]%
        {crainic1997planning}
\bibfield{author}{\bibinfo{person}{Teodor~Gabriel Crainic} {and}
  \bibinfo{person}{Gilbert Laporte}.} \bibinfo{year}{1997}\natexlab{}.
\newblock \showarticletitle{Planning models for freight transportation}.
\newblock \bibinfo{journal}{\emph{European journal of operational research}}
  \bibinfo{volume}{97}, \bibinfo{number}{3} (\bibinfo{year}{1997}),
  \bibinfo{pages}{409--438}.
\newblock


\bibitem[Dosovitskiy et~al\mbox{.}(2021)]%
        {DBLP:conf/iclr/DosovitskiyB0WZ21}
\bibfield{author}{\bibinfo{person}{Alexey Dosovitskiy}, \bibinfo{person}{Lucas
  Beyer}, \bibinfo{person}{Alexander Kolesnikov}, \bibinfo{person}{Dirk
  Weissenborn}, \bibinfo{person}{Xiaohua Zhai}, \bibinfo{person}{Thomas
  Unterthiner}, \bibinfo{person}{Mostafa Dehghani}, \bibinfo{person}{Matthias
  Minderer}, \bibinfo{person}{Georg Heigold}, \bibinfo{person}{Sylvain Gelly},
  \bibinfo{person}{Jakob Uszkoreit}, {and} \bibinfo{person}{Neil Houlsby}.}
  \bibinfo{year}{2021}\natexlab{}.
\newblock \showarticletitle{An Image is Worth 16x16 Words: Transformers for
  Image Recognition at Scale}. In \bibinfo{booktitle}{\emph{ICLR}}.
\newblock


\bibitem[Feng et~al\mbox{.}(2018)]%
        {feng2018deepmove}
\bibfield{author}{\bibinfo{person}{Jie Feng}, \bibinfo{person}{Yong Li},
  \bibinfo{person}{Chao Zhang}, \bibinfo{person}{Funing Sun},
  \bibinfo{person}{Fanchao Meng}, \bibinfo{person}{Ang Guo}, {and}
  \bibinfo{person}{Depeng Jin}.} \bibinfo{year}{2018}\natexlab{}.
\newblock \showarticletitle{Deepmove: Predicting human mobility with
  attentional recurrent networks}. In \bibinfo{booktitle}{\emph{WWW}}.
  \bibinfo{pages}{1459--1468}.
\newblock


\bibitem[Fu et~al\mbox{.}(2020)]%
        {fu2020compacteta}
\bibfield{author}{\bibinfo{person}{Kun Fu}, \bibinfo{person}{Fanlin Meng},
  \bibinfo{person}{Jieping Ye}, {and} \bibinfo{person}{Zheng Wang}.}
  \bibinfo{year}{2020}\natexlab{}.
\newblock \showarticletitle{Compacteta: A fast inference system for travel time
  prediction}. In \bibinfo{booktitle}{\emph{ACM SIGKDD}}.
  \bibinfo{pages}{3337--3345}.
\newblock


\bibitem[Gan et~al\mbox{.}(2021)]%
        {gan2021travel}
\bibfield{author}{\bibinfo{person}{Yunchong Gan}, \bibinfo{person}{Haoyu
  Zhang}, {and} \bibinfo{person}{Mingjie Wang}.}
  \bibinfo{year}{2021}\natexlab{}.
\newblock \showarticletitle{Travel Time Estimation Based on Neural Network with
  Auxiliary Loss}. In \bibinfo{booktitle}{\emph{SIGSPATIAL}}.
  \bibinfo{pages}{642--645}.
\newblock


\bibitem[Guo et~al\mbox{.}(2020)]%
        {guo2020attentional}
\bibfield{author}{\bibinfo{person}{Qing Guo}, \bibinfo{person}{Zhu Sun},
  \bibinfo{person}{Jie Zhang}, {and} \bibinfo{person}{Yin-Leng Theng}.}
  \bibinfo{year}{2020}\natexlab{}.
\newblock \showarticletitle{An attentional recurrent neural network for
  personalized next location recommendation}. In
  \bibinfo{booktitle}{\emph{AAAI}}, Vol.~\bibinfo{volume}{34}.
  \bibinfo{pages}{83--90}.
\newblock


\bibitem[Han et~al\mbox{.}(2022)]%
        {han2022deeptea}
\bibfield{author}{\bibinfo{person}{Xiaolin Han}, \bibinfo{person}{Reynold
  Cheng}, \bibinfo{person}{Chenhao Ma}, {and} \bibinfo{person}{Tobias
  Grubenmann}.} \bibinfo{year}{2022}\natexlab{}.
\newblock \showarticletitle{DeepTEA: effective and efficient online
  time-dependent trajectory outlier detection}.
\newblock \bibinfo{journal}{\emph{Proc. VLDB Endow.}} \bibinfo{volume}{15},
  \bibinfo{number}{7} (\bibinfo{year}{2022}), \bibinfo{pages}{1493--1505}.
\newblock


\bibitem[Hendrycks and Gimpel(2016)]%
        {hendrycks2016gaussian}
\bibfield{author}{\bibinfo{person}{Dan Hendrycks} {and} \bibinfo{person}{Kevin
  Gimpel}.} \bibinfo{year}{2016}\natexlab{}.
\newblock \showarticletitle{Gaussian error linear units (gelus)}.
\newblock \bibinfo{journal}{\emph{arXiv preprint arXiv:1606.08415}}
  (\bibinfo{year}{2016}).
\newblock


\bibitem[Ho et~al\mbox{.}(2020)]%
        {ho2020denoising}
\bibfield{author}{\bibinfo{person}{Jonathan Ho}, \bibinfo{person}{Ajay Jain},
  {and} \bibinfo{person}{Pieter Abbeel}.} \bibinfo{year}{2020}\natexlab{}.
\newblock \showarticletitle{Denoising diffusion probabilistic models}. In
  \bibinfo{booktitle}{\emph{NeurIPS}}, Vol.~\bibinfo{volume}{33}.
  \bibinfo{pages}{6840--6851}.
\newblock


\bibitem[Hochreiter and Schmidhuber(1997)]%
        {hochreiter1997long}
\bibfield{author}{\bibinfo{person}{Sepp Hochreiter} {and}
  \bibinfo{person}{J{\"u}rgen Schmidhuber}.} \bibinfo{year}{1997}\natexlab{}.
\newblock \showarticletitle{Long short-term memory}.
\newblock \bibinfo{journal}{\emph{Neural computation}} \bibinfo{volume}{9},
  \bibinfo{number}{8} (\bibinfo{year}{1997}), \bibinfo{pages}{1735--1780}.
\newblock


\bibitem[Huang et~al\mbox{.}(2021)]%
        {huang2021learning}
\bibfield{author}{\bibinfo{person}{Shuai Huang}, \bibinfo{person}{Yong Wang},
  \bibinfo{person}{Tianyu Zhao}, {and} \bibinfo{person}{Guoliang Li}.}
  \bibinfo{year}{2021}\natexlab{}.
\newblock \showarticletitle{A learning-based method for computing shortest path
  distances on road networks}. In \bibinfo{booktitle}{\emph{ICDE}}.
  \bibinfo{pages}{360--371}.
\newblock


\bibitem[Id{\'e} and Sugiyama(2011)]%
        {ide2011trajectory}
\bibfield{author}{\bibinfo{person}{Tsuyoshi Id{\'e}} {and}
  \bibinfo{person}{Masashi Sugiyama}.} \bibinfo{year}{2011}\natexlab{}.
\newblock \showarticletitle{Trajectory regression on road networks}. In
  \bibinfo{booktitle}{\emph{AAAI}}, Vol.~\bibinfo{volume}{25}.
  \bibinfo{pages}{203--208}.
\newblock


\bibitem[Jarzynski(1997)]%
        {jarzynski1997equilibrium}
\bibfield{author}{\bibinfo{person}{Christopher Jarzynski}.}
  \bibinfo{year}{1997}\natexlab{}.
\newblock \showarticletitle{Equilibrium free-energy differences from
  nonequilibrium measurements: A master-equation approach}.
\newblock \bibinfo{journal}{\emph{Physical Review E}} \bibinfo{volume}{56},
  \bibinfo{number}{5} (\bibinfo{year}{1997}), \bibinfo{pages}{5018}.
\newblock


\bibitem[Jensen and Tradi{\v{s}}auskas(2009)]%
        {Jensen2009}
\bibfield{author}{\bibinfo{person}{Christian~S. Jensen} {and}
  \bibinfo{person}{Nerius Tradi{\v{s}}auskas}.}
  \bibinfo{year}{2009}\natexlab{}.
\newblock \showarticletitle{Map Matching}. In
  \bibinfo{booktitle}{\emph{Encyclopedia of Database Systems}}.
  \bibinfo{pages}{1692--1696}.
\newblock


\bibitem[Jin et~al\mbox{.}(2021)]%
        {jin2021hierarchical}
\bibfield{author}{\bibinfo{person}{Guangyin Jin}, \bibinfo{person}{Huan Yan},
  \bibinfo{person}{Fuxian Li}, \bibinfo{person}{Yong Li}, {and}
  \bibinfo{person}{Jincai Huang}.} \bibinfo{year}{2021}\natexlab{}.
\newblock \showarticletitle{Hierarchical neural architecture search for travel
  time estimation}. In \bibinfo{booktitle}{\emph{SIGSPATIAL}}.
  \bibinfo{pages}{91--94}.
\newblock


\bibitem[Jindal et~al\mbox{.}(2017)]%
        {jindal2017unified}
\bibfield{author}{\bibinfo{person}{Ishan Jindal}, \bibinfo{person}{Xuewen
  Chen}, \bibinfo{person}{Matthew Nokleby}, \bibinfo{person}{Jieping Ye},
  {et~al\mbox{.}}} \bibinfo{year}{2017}\natexlab{}.
\newblock \showarticletitle{A unified neural network approach for estimating
  travel time and distance for a taxi trip}.
\newblock \bibinfo{journal}{\emph{arXiv preprint arXiv:1710.04350}}
  (\bibinfo{year}{2017}).
\newblock


\bibitem[Johnson(1973)]%
        {johnson1973note}
\bibfield{author}{\bibinfo{person}{Donald~B Johnson}.}
  \bibinfo{year}{1973}\natexlab{}.
\newblock \showarticletitle{A note on Dijkstra's shortest path algorithm}.
\newblock \bibinfo{journal}{\emph{J. ACM}} \bibinfo{volume}{20},
  \bibinfo{number}{3} (\bibinfo{year}{1973}), \bibinfo{pages}{385--388}.
\newblock


\bibitem[Ke et~al\mbox{.}(2017)]%
        {ke2017lightgbm}
\bibfield{author}{\bibinfo{person}{Guolin Ke}, \bibinfo{person}{Qi Meng},
  \bibinfo{person}{Thomas Finley}, \bibinfo{person}{Taifeng Wang},
  \bibinfo{person}{Wei Chen}, \bibinfo{person}{Weidong Ma},
  \bibinfo{person}{Qiwei Ye}, {and} \bibinfo{person}{Tie-Yan Liu}.}
  \bibinfo{year}{2017}\natexlab{}.
\newblock \showarticletitle{Lightgbm: A highly efficient gradient boosting
  decision tree}. In \bibinfo{booktitle}{\emph{NeurIPS}},
  Vol.~\bibinfo{volume}{30}. \bibinfo{pages}{3146--3154}.
\newblock


\bibitem[Kong et~al\mbox{.}(2020)]%
        {kong2020diffwave}
\bibfield{author}{\bibinfo{person}{Zhifeng Kong}, \bibinfo{person}{Wei Ping},
  \bibinfo{person}{Jiaji Huang}, \bibinfo{person}{Kexin Zhao}, {and}
  \bibinfo{person}{Bryan Catanzaro}.} \bibinfo{year}{2020}\natexlab{}.
\newblock \showarticletitle{Diffwave: A versatile diffusion model for audio
  synthesis}.
\newblock \bibinfo{journal}{\emph{arXiv preprint arXiv:2009.09761}}
  (\bibinfo{year}{2020}).
\newblock


\bibitem[Li et~al\mbox{.}(2020)]%
        {li2020spatial}
\bibfield{author}{\bibinfo{person}{Xiucheng Li}, \bibinfo{person}{Gao Cong},
  {and} \bibinfo{person}{Yun Cheng}.} \bibinfo{year}{2020}\natexlab{}.
\newblock \showarticletitle{Spatial transition learning on road networks with
  deep probabilistic models}. In \bibinfo{booktitle}{\emph{ICDE}}.
  \bibinfo{pages}{349--360}.
\newblock


\bibitem[Li et~al\mbox{.}(2019)]%
        {li2019learning}
\bibfield{author}{\bibinfo{person}{Xiucheng Li}, \bibinfo{person}{Gao Cong},
  \bibinfo{person}{Aixin Sun}, {and} \bibinfo{person}{Yun Cheng}.}
  \bibinfo{year}{2019}\natexlab{}.
\newblock \showarticletitle{Learning travel time distributions with deep
  generative model}. In \bibinfo{booktitle}{\emph{WWW}}.
  \bibinfo{pages}{1017--1027}.
\newblock


\bibitem[Li et~al\mbox{.}(2015)]%
        {li2015towards}
\bibfield{author}{\bibinfo{person}{Yaguang Li}, \bibinfo{person}{Dingxiong
  Deng}, \bibinfo{person}{Ugur Demiryurek}, \bibinfo{person}{Cyrus Shahabi},
  {and} \bibinfo{person}{Siva Ravada}.} \bibinfo{year}{2015}\natexlab{}.
\newblock \showarticletitle{Towards fast and accurate solutions to vehicle
  routing in a large-scale and dynamic environment}. In
  \bibinfo{booktitle}{\emph{SSTD}}. \bibinfo{pages}{119--136}.
\newblock


\bibitem[Li et~al\mbox{.}(2018)]%
        {li2018multi}
\bibfield{author}{\bibinfo{person}{Yaguang Li}, \bibinfo{person}{Kun Fu},
  \bibinfo{person}{Zheng Wang}, \bibinfo{person}{Cyrus Shahabi},
  \bibinfo{person}{Jieping Ye}, {and} \bibinfo{person}{Yan Liu}.}
  \bibinfo{year}{2018}\natexlab{}.
\newblock \showarticletitle{Multi-task representation learning for travel time
  estimation}. In \bibinfo{booktitle}{\emph{ACM SIGKDD}}.
  \bibinfo{pages}{1695--1704}.
\newblock


\bibitem[Lin et~al\mbox{.}(2021)]%
        {lin2021pre}
\bibfield{author}{\bibinfo{person}{Yan Lin}, \bibinfo{person}{Huaiyu Wan},
  \bibinfo{person}{Shengnan Guo}, {and} \bibinfo{person}{Youfang Lin}.}
  \bibinfo{year}{2021}\natexlab{}.
\newblock \showarticletitle{Pre-training context and time aware location
  embeddings from spatial-temporal trajectories for user next location
  prediction}. In \bibinfo{booktitle}{\emph{AAAI}}, Vol.~\bibinfo{volume}{35}.
  \bibinfo{pages}{4241--4248}.
\newblock


\bibitem[Litman(2009)]%
        {litman2009transportation}
\bibfield{author}{\bibinfo{person}{Todd Litman}.}
  \bibinfo{year}{2009}\natexlab{}.
\newblock \showarticletitle{Transportation cost and benefit analysis}.
\newblock \bibinfo{journal}{\emph{Victoria Transport Policy Institute}}
  \bibinfo{volume}{31} (\bibinfo{year}{2009}), \bibinfo{pages}{1--19}.
\newblock


\bibitem[Liu et~al\mbox{.}(2017)]%
        {liu2017finding}
\bibfield{author}{\bibinfo{person}{Huiping Liu}, \bibinfo{person}{Cheqing Jin},
  \bibinfo{person}{Bin Yang}, {and} \bibinfo{person}{Aoying Zhou}.}
  \bibinfo{year}{2017}\natexlab{}.
\newblock \showarticletitle{Finding top-k shortest paths with diversity}.
\newblock \bibinfo{journal}{\emph{IEEE Trans. on Know. and Data Eng.}}
  \bibinfo{volume}{30}, \bibinfo{number}{3} (\bibinfo{year}{2017}),
  \bibinfo{pages}{488--502}.
\newblock


\bibitem[Liu et~al\mbox{.}(2022)]%
        {liu2022convnet}
\bibfield{author}{\bibinfo{person}{Zhuang Liu}, \bibinfo{person}{Hanzi Mao},
  \bibinfo{person}{Chao-Yuan Wu}, \bibinfo{person}{Christoph Feichtenhofer},
  \bibinfo{person}{Trevor Darrell}, {and} \bibinfo{person}{Saining Xie}.}
  \bibinfo{year}{2022}\natexlab{}.
\newblock \showarticletitle{A convnet for the 2020s}. In
  \bibinfo{booktitle}{\emph{IEEE CVPR}}. \bibinfo{pages}{11976--11986}.
\newblock


\bibitem[Nelson(1998)]%
        {nelson1998time}
\bibfield{author}{\bibinfo{person}{Brian~K Nelson}.}
  \bibinfo{year}{1998}\natexlab{}.
\newblock \showarticletitle{Time series analysis using autoregressive
  integrated moving average (ARIMA) models}.
\newblock \bibinfo{journal}{\emph{Academic emergency medicine}}
  \bibinfo{volume}{5}, \bibinfo{number}{7} (\bibinfo{year}{1998}),
  \bibinfo{pages}{739--744}.
\newblock


\bibitem[Paszke et~al\mbox{.}(2019)]%
        {paszke2019pytorch}
\bibfield{author}{\bibinfo{person}{Adam Paszke}, \bibinfo{person}{Sam Gross},
  \bibinfo{person}{Francisco Massa}, \bibinfo{person}{Adam Lerer},
  \bibinfo{person}{James Bradbury}, \bibinfo{person}{Gregory Chanan},
  \bibinfo{person}{Trevor Killeen}, \bibinfo{person}{Zeming Lin},
  \bibinfo{person}{Natalia Gimelshein}, \bibinfo{person}{Luca Antiga},
  {et~al\mbox{.}}} \bibinfo{year}{2019}\natexlab{}.
\newblock \showarticletitle{PyTorch: An imperative style, high-performance deep
  learning library}. In \bibinfo{booktitle}{\emph{NeurIPS}}.
  \bibinfo{pages}{8024--8035}.
\newblock


\bibitem[Rombach et~al\mbox{.}(2022)]%
        {rombach2022high}
\bibfield{author}{\bibinfo{person}{Robin Rombach}, \bibinfo{person}{Andreas
  Blattmann}, \bibinfo{person}{Dominik Lorenz}, \bibinfo{person}{Patrick
  Esser}, {and} \bibinfo{person}{Bj{\"o}rn Ommer}.}
  \bibinfo{year}{2022}\natexlab{}.
\newblock \showarticletitle{High-resolution image synthesis with latent
  diffusion models}. In \bibinfo{booktitle}{\emph{IEEE CVPR}}.
  \bibinfo{pages}{10684--10695}.
\newblock


\bibitem[Ronneberger et~al\mbox{.}(2015)]%
        {ronneberger2015u}
\bibfield{author}{\bibinfo{person}{Olaf Ronneberger}, \bibinfo{person}{Philipp
  Fischer}, {and} \bibinfo{person}{Thomas Brox}.}
  \bibinfo{year}{2015}\natexlab{}.
\newblock \showarticletitle{U-net: Convolutional networks for biomedical image
  segmentation}. In \bibinfo{booktitle}{\emph{MICCAI}}.
  \bibinfo{pages}{234--241}.
\newblock


\bibitem[Ruan et~al\mbox{.}(2020)]%
        {ruan2020doing}
\bibfield{author}{\bibinfo{person}{Sijie Ruan}, \bibinfo{person}{Zi Xiong},
  \bibinfo{person}{Cheng Long}, \bibinfo{person}{Yiheng Chen},
  \bibinfo{person}{Jie Bao}, \bibinfo{person}{Tianfu He},
  \bibinfo{person}{Ruiyuan Li}, \bibinfo{person}{Shengnan Wu},
  \bibinfo{person}{Zhongyuan Jiang}, {and} \bibinfo{person}{Yu Zheng}.}
  \bibinfo{year}{2020}\natexlab{}.
\newblock \showarticletitle{Doing in One Go: Delivery Time Inference Based on
  Couriers' Trajectories}. In \bibinfo{booktitle}{\emph{ACM SIGKDD}}.
  \bibinfo{pages}{2813--2821}.
\newblock


\bibitem[Sankaranarayanan and Samet(2009)]%
        {dist_oracle}
\bibfield{author}{\bibinfo{person}{Jagan Sankaranarayanan} {and}
  \bibinfo{person}{Hanan Samet}.} \bibinfo{year}{2009}\natexlab{}.
\newblock \showarticletitle{Distance Oracles for Spatial Networks}. In
  \bibinfo{booktitle}{\emph{ICDE}}. \bibinfo{pages}{652–663}.
\newblock


\bibitem[Sankaranarayanan et~al\mbox{.}(2009)]%
        {path_oracle}
\bibfield{author}{\bibinfo{person}{Jagan Sankaranarayanan},
  \bibinfo{person}{Hanan Samet}, {and} \bibinfo{person}{Houman Alborzi}.}
  \bibinfo{year}{2009}\natexlab{}.
\newblock \showarticletitle{Path Oracles for Spatial Networks}.
\newblock \bibinfo{journal}{\emph{Proc. VLDB Endow.}} \bibinfo{volume}{2},
  \bibinfo{number}{1} (\bibinfo{year}{2009}), \bibinfo{pages}{1210–1221}.
\newblock


\bibitem[Sohl-Dickstein et~al\mbox{.}(2015)]%
        {sohl2015deep}
\bibfield{author}{\bibinfo{person}{Jascha Sohl-Dickstein},
  \bibinfo{person}{Eric Weiss}, \bibinfo{person}{Niru Maheswaranathan}, {and}
  \bibinfo{person}{Surya Ganguli}.} \bibinfo{year}{2015}\natexlab{}.
\newblock \showarticletitle{Deep unsupervised learning using nonequilibrium
  thermodynamics}. In \bibinfo{booktitle}{\emph{ICML}}.
  \bibinfo{pages}{2256--2265}.
\newblock


\bibitem[Song and Ermon(2019)]%
        {song2019generative}
\bibfield{author}{\bibinfo{person}{Yang Song} {and} \bibinfo{person}{Stefano
  Ermon}.} \bibinfo{year}{2019}\natexlab{}.
\newblock \showarticletitle{Generative modeling by estimating gradients of the
  data distribution}. In \bibinfo{booktitle}{\emph{NeurIPS}},
  Vol.~\bibinfo{volume}{32}. \bibinfo{pages}{11895--11907}.
\newblock


\bibitem[Sun et~al\mbox{.}(2022)]%
        {sun2022deep}
\bibfield{author}{\bibinfo{person}{Fuyong Sun}, \bibinfo{person}{Ruipeng Gao},
  \bibinfo{person}{Weiwei Xing}, \bibinfo{person}{Yaoxue Zhang},
  \bibinfo{person}{Wei Lu}, \bibinfo{person}{Jun Fang}, {and}
  \bibinfo{person}{Shui Liu}.} \bibinfo{year}{2022}\natexlab{}.
\newblock \showarticletitle{Deep Fusion for Travel Time Estimation Based on
  Road Network Topology}.
\newblock \bibinfo{journal}{\emph{IEEE Intelligent Systems}}
  \bibinfo{volume}{37}, \bibinfo{number}{3} (\bibinfo{year}{2022}),
  \bibinfo{pages}{98--107}.
\newblock


\bibitem[Tashiro et~al\mbox{.}(2021)]%
        {tashiro2021csdi}
\bibfield{author}{\bibinfo{person}{Yusuke Tashiro}, \bibinfo{person}{Jiaming
  Song}, \bibinfo{person}{Yang Song}, {and} \bibinfo{person}{Stefano Ermon}.}
  \bibinfo{year}{2021}\natexlab{}.
\newblock \showarticletitle{CSDI: Conditional score-based diffusion models for
  probabilistic time series imputation}. In
  \bibinfo{booktitle}{\emph{NeurIPS}}, Vol.~\bibinfo{volume}{34}.
  \bibinfo{pages}{24804--24816}.
\newblock


\bibitem[Vaswani et~al\mbox{.}(2017)]%
        {vaswani2017attention}
\bibfield{author}{\bibinfo{person}{Ashish Vaswani}, \bibinfo{person}{Noam
  Shazeer}, \bibinfo{person}{Niki Parmar}, \bibinfo{person}{Jakob Uszkoreit},
  \bibinfo{person}{Llion Jones}, \bibinfo{person}{Aidan~N Gomez},
  \bibinfo{person}{{\L}ukasz Kaiser}, {and} \bibinfo{person}{Illia
  Polosukhin}.} \bibinfo{year}{2017}\natexlab{}.
\newblock \showarticletitle{Attention is all you need}. In
  \bibinfo{booktitle}{\emph{NeurIPS}}, Vol.~\bibinfo{volume}{30}.
  \bibinfo{pages}{5998--6008}.
\newblock


\bibitem[Wang et~al\mbox{.}(2022)]%
        {wang2022fine}
\bibfield{author}{\bibinfo{person}{Chenxing Wang}, \bibinfo{person}{Fang Zhao},
  \bibinfo{person}{Haichao Zhang}, \bibinfo{person}{Haiyong Luo},
  \bibinfo{person}{Yanjun Qin}, {and} \bibinfo{person}{Yuchen Fang}.}
  \bibinfo{year}{2022}\natexlab{}.
\newblock \showarticletitle{Fine-Grained Trajectory-Based Travel Time
  Estimation for Multi-City Scenarios Based on Deep Meta-Learning}.
\newblock \bibinfo{journal}{\emph{IEEE Trans. on Intelli. Trans. Sys.}}
  \bibinfo{volume}{23}, \bibinfo{number}{9} (\bibinfo{year}{2022}),
  \bibinfo{pages}{15716--15728}.
\newblock


\bibitem[Wang et~al\mbox{.}(2018b)]%
        {wang2018will}
\bibfield{author}{\bibinfo{person}{Dong Wang}, \bibinfo{person}{Junbo Zhang},
  \bibinfo{person}{Wei Cao}, \bibinfo{person}{Jian Li}, {and}
  \bibinfo{person}{Yu Zheng}.} \bibinfo{year}{2018}\natexlab{b}.
\newblock \showarticletitle{When will you arrive? Estimating travel time based
  on deep neural networks}. In \bibinfo{booktitle}{\emph{AAAI}},
  Vol.~\bibinfo{volume}{32}. \bibinfo{pages}{2500--2507}.
\newblock


\bibitem[Wang et~al\mbox{.}(2019)]%
        {wang2019simple}
\bibfield{author}{\bibinfo{person}{Hongjian Wang}, \bibinfo{person}{Xianfeng
  Tang}, \bibinfo{person}{Yu-Hsuan Kuo}, \bibinfo{person}{Daniel Kifer}, {and}
  \bibinfo{person}{Zhenhui Li}.} \bibinfo{year}{2019}\natexlab{}.
\newblock \showarticletitle{A simple baseline for travel time estimation using
  large-scale trip data}.
\newblock \bibinfo{journal}{\emph{ACM Trans. on Intelli. Sys. and Tech.}}
  \bibinfo{volume}{10}, \bibinfo{number}{2} (\bibinfo{year}{2019}),
  \bibinfo{pages}{1--22}.
\newblock


\bibitem[Wang et~al\mbox{.}(2021)]%
        {wang2021passenger}
\bibfield{author}{\bibinfo{person}{Yuandong Wang}, \bibinfo{person}{Hongzhi
  Yin}, \bibinfo{person}{Tong Chen}, \bibinfo{person}{Chunyang Liu},
  \bibinfo{person}{Ben Wang}, \bibinfo{person}{Tianyu Wo}, {and}
  \bibinfo{person}{Jie Xu}.} \bibinfo{year}{2021}\natexlab{}.
\newblock \showarticletitle{Passenger Mobility Prediction via Representation
  Learning for Dynamic Directed and Weighted Graphs}.
\newblock \bibinfo{journal}{\emph{ACM Trans. on Intelli. Sys. and Tech.}}
  \bibinfo{volume}{13}, \bibinfo{number}{1} (\bibinfo{year}{2021}),
  \bibinfo{pages}{1--25}.
\newblock


\bibitem[Wang et~al\mbox{.}(2014)]%
        {wang2014travel}
\bibfield{author}{\bibinfo{person}{Yilun Wang}, \bibinfo{person}{Yu Zheng},
  {and} \bibinfo{person}{Yexiang Xue}.} \bibinfo{year}{2014}\natexlab{}.
\newblock \showarticletitle{Travel time estimation of a path using sparse
  trajectories}. In \bibinfo{booktitle}{\emph{ACM SIGKDD}}.
  \bibinfo{pages}{25--34}.
\newblock


\bibitem[Wang et~al\mbox{.}(2018a)]%
        {wang2018learning}
\bibfield{author}{\bibinfo{person}{Zheng Wang}, \bibinfo{person}{Kun Fu}, {and}
  \bibinfo{person}{Jieping Ye}.} \bibinfo{year}{2018}\natexlab{a}.
\newblock \showarticletitle{Learning to estimate the travel time}. In
  \bibinfo{booktitle}{\emph{ACM SIGKDD}}. \bibinfo{pages}{858--866}.
\newblock


\bibitem[Wen et~al\mbox{.}(2021)]%
        {wen2021package}
\bibfield{author}{\bibinfo{person}{Haomin Wen}, \bibinfo{person}{Youfang Lin},
  \bibinfo{person}{Fan Wu}, \bibinfo{person}{Huaiyu Wan},
  \bibinfo{person}{Shengnan Guo}, \bibinfo{person}{Lixia Wu},
  \bibinfo{person}{Chao Song}, {and} \bibinfo{person}{Yinghui Xu}.}
  \bibinfo{year}{2021}\natexlab{}.
\newblock \showarticletitle{Package pick-up route prediction via modeling
  couriers’ spatial-temporal behaviors}. In \bibinfo{booktitle}{\emph{IEEE
  ICDE}}. \bibinfo{pages}{2141--2146}.
\newblock


\bibitem[Wu and Wu(2019)]%
        {wu2019deepeta}
\bibfield{author}{\bibinfo{person}{Fan Wu} {and} \bibinfo{person}{Lixia Wu}.}
  \bibinfo{year}{2019}\natexlab{}.
\newblock \showarticletitle{DeepETA: a spatial-temporal sequential neural
  network model for estimating time of arrival in package delivery system}. In
  \bibinfo{booktitle}{\emph{AAAI}}, Vol.~\bibinfo{volume}{33}.
  \bibinfo{pages}{774--781}.
\newblock


\bibitem[Wu et~al\mbox{.}(2016)]%
        {wu2016probabilistic}
\bibfield{author}{\bibinfo{person}{Hao Wu}, \bibinfo{person}{Jiangyun Mao},
  \bibinfo{person}{Weiwei Sun}, \bibinfo{person}{Baihua Zheng},
  \bibinfo{person}{Hanyuan Zhang}, \bibinfo{person}{Ziyang Chen}, {and}
  \bibinfo{person}{Wei Wang}.} \bibinfo{year}{2016}\natexlab{}.
\newblock \showarticletitle{Probabilistic robust route recovery with
  spatio-temporal dynamics}. In \bibinfo{booktitle}{\emph{SIGKDD}}.
  \bibinfo{pages}{1915--1924}.
\newblock


\bibitem[Xu et~al\mbox{.}(2021)]%
        {xu2021taml}
\bibfield{author}{\bibinfo{person}{Jiajie Xu}, \bibinfo{person}{Saijun Xu},
  \bibinfo{person}{Rui Zhou}, \bibinfo{person}{Chengfei Liu},
  \bibinfo{person}{An Liu}, {and} \bibinfo{person}{Lei Zhao}.}
  \bibinfo{year}{2021}\natexlab{}.
\newblock \showarticletitle{TAML: A Traffic-aware Multi-task Learning Model for
  Estimating Travel Time}.
\newblock \bibinfo{journal}{\emph{ACM Trans. on Intelli. Sys. and Tech.}}
  \bibinfo{volume}{12}, \bibinfo{number}{6} (\bibinfo{year}{2021}),
  \bibinfo{pages}{1--14}.
\newblock


\bibitem[Xu et~al\mbox{.}(2020)]%
        {xu2020tadnm}
\bibfield{author}{\bibinfo{person}{Saijun Xu}, \bibinfo{person}{Jiajie Xu},
  \bibinfo{person}{Rui Zhou}, \bibinfo{person}{Chengfei Liu},
  \bibinfo{person}{Zhixu Li}, {and} \bibinfo{person}{An Liu}.}
  \bibinfo{year}{2020}\natexlab{}.
\newblock \showarticletitle{Tadnm: A transportation-mode aware deep neural
  model for travel time estimation}. In \bibinfo{booktitle}{\emph{DASFAA}}.
  \bibinfo{pages}{468--484}.
\newblock


\bibitem[Yang et~al\mbox{.}(2022)]%
        {yang2022multitask}
\bibfield{author}{\bibinfo{person}{Ling Yang}, \bibinfo{person}{Shouxu Jiang},
  {and} \bibinfo{person}{Fusheng Zhang}.} \bibinfo{year}{2022}\natexlab{}.
\newblock \showarticletitle{Multitask Learning with Graph Neural Network for
  Travel Time Estimation}.
\newblock \bibinfo{journal}{\emph{Computational Intelligence and Neuroscience}}
   \bibinfo{volume}{2022} (\bibinfo{year}{2022}).
\newblock


\bibitem[Yuan et~al\mbox{.}(2020)]%
        {yuan2020effective}
\bibfield{author}{\bibinfo{person}{Haitao Yuan}, \bibinfo{person}{Guoliang Li},
  \bibinfo{person}{Zhifeng Bao}, {and} \bibinfo{person}{Ling Feng}.}
  \bibinfo{year}{2020}\natexlab{}.
\newblock \showarticletitle{Effective travel time estimation: When historical
  trajectories over road networks matter}. In
  \bibinfo{booktitle}{\emph{SIGMOD}}. \bibinfo{pages}{2135--2149}.
\newblock


\bibitem[Zhang et~al\mbox{.}(2019)]%
        {zhang2019flow}
\bibfield{author}{\bibinfo{person}{Junbo Zhang}, \bibinfo{person}{Yu Zheng},
  \bibinfo{person}{Junkai Sun}, {and} \bibinfo{person}{Dekang Qi}.}
  \bibinfo{year}{2019}\natexlab{}.
\newblock \showarticletitle{Flow prediction in spatio-temporal networks based
  on multitask deep learning}.
\newblock \bibinfo{journal}{\emph{IEEE Trans. on Know. and Data Eng.}}
  \bibinfo{volume}{32}, \bibinfo{number}{3} (\bibinfo{year}{2019}),
  \bibinfo{pages}{468--478}.
\newblock


\bibitem[Zhou et~al\mbox{.}(2021b)]%
        {zhou2021informer}
\bibfield{author}{\bibinfo{person}{Haoyi Zhou}, \bibinfo{person}{Shanghang
  Zhang}, \bibinfo{person}{Jieqi Peng}, \bibinfo{person}{Shuai Zhang},
  \bibinfo{person}{Jianxin Li}, \bibinfo{person}{Hui Xiong}, {and}
  \bibinfo{person}{Wancai Zhang}.} \bibinfo{year}{2021}\natexlab{b}.
\newblock \showarticletitle{Informer: Beyond efficient transformer for long
  sequence time-series forecasting}. In \bibinfo{booktitle}{\emph{AAAI}},
  Vol.~\bibinfo{volume}{35}. \bibinfo{pages}{11106--11115}.
\newblock


\bibitem[Zhou et~al\mbox{.}(2021a)]%
        {zhou20213d}
\bibfield{author}{\bibinfo{person}{Linqi Zhou}, \bibinfo{person}{Yilun Du},
  {and} \bibinfo{person}{Jiajun Wu}.} \bibinfo{year}{2021}\natexlab{a}.
\newblock \showarticletitle{3d shape generation and completion through
  point-voxel diffusion}. In \bibinfo{booktitle}{\emph{IEEE ICCV}}.
  \bibinfo{pages}{5826--5835}.
\newblock


\end{thebibliography}

\end{document}